\documentclass[review]{elsarticle}

\usepackage{lineno, hyperref}
\modulolinenumbers[5]

\journal{Journal of \LaTeX\ Templates}

\usepackage{amsmath}
\usepackage{amssymb}
\usepackage{xcolor}
\usepackage{url}
\usepackage{subcaption}
\usepackage{tabularx}

\newcolumntype{Y}{>{\centering\arraybackslash}X}

\begin{document}
\begin{frontmatter}

\title{BabyNet: Reconstructing 3D faces of babies from uncalibrated photographs}


\author[upf]{Araceli Morales}

\author[antonio1,antonio2,antonio3]{Antonio R. Porras}

\author[marius1,marius2]{Marius George Linguraru}
\author[upf]{Gemma Piella}
\author[upf]{Federico M. Sukno\corref{mycorrespondingauthor}} 
\cortext[mycorrespondingauthor]{Corresponding author} \ead{federico.sukno@upf.com}

\address[upf]{Department of Information and Communication Technologies, Pompeu Fabra University, Barcelona, Spain.}
\address[antonio1]{Department of Biostatistics and Informatics, Colorado School of Public Health}
\address[antonio2]{Department of Pediatrics, School of Medicine, University of Colorado Anschutz Medical Campus, Aurora, CO, USA}
\address[antonio3]{Divisions of Pediatric Plastic \& Reconstructive Surgery and Neurosurgery, Children's Hospital Colorado, Aurora, CO, USA.}

\address[marius1]{Sheikh Zayed Institute for Pediatric Surgical Innovation, Children's National Hospital, Washington, DC, USA}
\address[marius2]{Departments of Radiology and Pediatrics from the George Washington University School of Medicine and Health Sciences, Washington, DC, USA}

\begin{abstract}
We present a 3D face reconstruction system that aims at recovering the 3D facial geometry of babies from uncalibrated photographs, BabyNet. Since the 3D facial geometry of babies differs substantially from that of adults, baby-specific facial reconstruction systems are needed. BabyNet consists of two stages: 1) a 3D graph convolutional autoencoder learns a latent space of the baby 3D facial shape; and 2) a 2D encoder that maps photographs to the 3D latent space based on representative features extracted using transfer learning. In this way, using the pre-trained 3D decoder, we can recover a 3D face from 2D images. We evaluate BabyNet and show that 1) methods based on adult datasets cannot model the 3D facial geometry of babies, which proves the need for a baby-specific method, and 2) BabyNet outperforms classical model-fitting methods even when a baby-specific 3D morphable model, such as BabyFM, is used.
\end{abstract}


\end{frontmatter}


\section{Introduction}

Three-dimensional (3D) face reconstruction from uncalibrated two-dimensional (2D) images is a long-standing problem in computer vision with many different applications, such as face recognition \citep{JZhaoIJCAI2018,FLiuICCV2019,JGuoCCBR2018,FLiuPAMI2020}, face alignment \citep{GZhangFG2018,XTuToM2020,XZhuPAMI2019,YFengECCV2018}, image edition \citep{JCaoNIPS2018,ZGengCVPR2019,JCaoIJCV2020}, face animation \citep{CCaoTOG2015,CCaoTOG2016,LHuTOG2017}, age estimation \citep{SavovICCVW2019}, or medical diagnostic purposes \citep{LTuMICCAI2018,LTuMICCAI2019,AlomarVISAPP2021}. Despite its great amount of advantages, 3D face reconstruction is a very challenging task since it is inherently ill-posed. Because of the 3D nature of the face, a 2D image is insufficient to accurately capture its geometry, as it collapses one dimension, and thus loses information regarding the actual 3D geometry. As a result, there are ambiguities in the solution because the same 2D projection can be generated from different 3D geometries. 

Although different strategies to address these ambiguities have been presented for many years, it was not until the introduction of 3D morphable models (3DMMs) \citep{BlanzVetterSIGGRAPH1999} that this field gained substantial popularity. 3DMMs are statistical models that encode the 3D geometric variations of the human face into a linear vector space. Hence, they consist of a mean 3D shape and a set of unit vectors that indicate the principal modes of variations. With this statistical representation of the 3D facial geometry, 3DMM can be used to regularise 3D face reconstruction methods by constraining the space of plausible solutions, i.e., solutions that look like real faces. Furthermore, 3DMMs can be used to reconstruct the 3D face from multiple or even a single 2D images by finding the best fitting linear combination of the model bases.

In the last few years, many deep learning-based methods have been proposed to solve the 3D face reconstruction problem. The power of these methods rely on their ability to learn non-linearities from the training data at the expense of needing large training sets. For the specific problem of 3D face reconstruction from uncalibrated 2D images, such training data consists in 2D pictures and their corresponding 3D facial geometries, whose availability is often limited. To mitigate this problem, different groups \citep{XZhuCVPR2016,XLiICASSP2020,Richardson3DV2016,GenovaCVPR2018} have proposed the generation of synthetic datasets based on 3DMM to train deep learning-based methods. However, given the linear nature of the 3DMMs, the synthetic data is not as detailed as desired. Moreover, deep learning methods require data samples with a common fixed topology, which requires homogeneous sizes of 2D images, and the same triangulation, number of vertices and anatomical correspondences between 3D pictures.

Although many different approaches have been proposed to tackle the 3D face reconstruction problem, none of them provides a solution for 3D face reconstruction of babies. This application is especially relevant to enable medical diagnosis based on cranio-facial imaging data \citep{LTuMICCAI2018,PorrasPlastRecSurg2019,ChendebEMBC2015}. A challenge is that the facial geometry of babies is very different from that older children or adults. Hence, a baby-specific
method to recover the 3D face of babies is needed.

In this work, we present the first 3D face reconstruction method that specifically targets babies: BabyNet. Our method is based on a two-phased deep neural network consisting of: 1) the estimation of a low-dimensional latent space of the 3D faces using a \textit{3D autoencoder}, and 2) training a \textit{2D encoder} to produce the same latent representation as the \emph{3D autoencoder} for 2D images, so that the \emph{3D decoder} can reconstruct the corresponding 3D face. We adopt an emerging type of networks called graph convolutional networks (GCNs) to design the \textit{3D autoencoder}, which extend convolutional neural networks to non-Euclidean domains, such as 3D surface meshes. Specifically, they allow directly processing the 3D faces without the need for alternative Euclidean representations, which usually negatively impact the reconstruction accuracy. Additionally, our \textit{2D encoder} extracts meaningful image features using transfer learning from a pre-trained network. We use both the predicted feature vector by its output layer and feature maps estimated by its intermediate layers to capture both global and detailed facial information, respectively. This approach allows us to estimate a 3D latent representation that is then decoded to reconstruct a 3D face by the \textit{3D decoder}.

Because of the low availability of data from babies to train our network, we used our recent Baby Face Model (BabyFM) \citep{MoralesFG2020,MoralesPAMI2022} – a 3DMM specific to babies – to construct a large enough training dataset. We evaluate our BabyNet by comparing its reconstruction ability with that of other state-of-the-art deep learning-based methods trained using adult data to demonstrate the need for a facial reconstruction method specific to babies. Furthermore, to allow for a more honest evaluation, we also compare our BabyNet with classical 3DMM fitting approaches, using as 3DMM the newly presented BabyFM \citep{MoralesFG2020,MoralesPAMI2022}.

The contributions of this paper can thus be summarised as follows:
\begin{enumerate}
    \item We present a 3D face reconstruction method that specifically targets babies, which we trained thanks to the BabyFM.
    \item We design an GCN autoencoder to learn a non-linear latent space of the 3D facial shape variations of babies that allows us to directly tackle the original non-Euclidean representation of 3D facial data, instead of using alternative intermediate representations.
    \item We apply transfer learning to take advantage of a powerful pre-trained image encoder. We use its output feature vector and its intermediate layer outputs to capture both global and fine facial details to create accurate 3D facial reconstructions.
\end{enumerate}

The rest of this manuscript is organised as follows. In Section \ref{sec:2RelatedWork}, we concisely review the related work, In Section \ref{sec:3BabyNet}, we present the details of the proposed architecture of BabyNet. In Section \ref{sec:4Training}, we describe the training procedure, and in Section \ref{sec:5SystemEvaluation}, we evaluate our major design choices. In Section \ref{sec:6CompWithSOTA}, we compare quantitatively and qualitatively the reconstruction accuracy of BabyNet to state-of-the-art methods evaluated on baby data. Finally, in Section \ref{sec:7Discussions}, we discuss the results and limitations and suggest open lines for future work, and, in Section \ref{sec:8Conclusions}, we provide some conclusions
\section{Related work} \label{sec:2RelatedWork}

In this section, we provide a global overview of the state-of-the-art on 3D face reconstruction methods, focusing on the works most related to our approach. For a comprehensive survey on the subject, we refer the reader to \cite{MoralesCompSciRev2021}.

\subsection{Traditional approaches}
The first attempts to reconstruct the 3D facial geometry from uncalibrated images were derived from classical photometric stereo methods \citep{WoodhamOpEng1980}. Under a fully controlled environment, these methods estimate the surface normals by observing the face under various lighting conditions. Although the resulting 3D faces were highly accurate, the need for controlled environments hinders their feasible implementation using real world data, so most later works focused on adapting this approach to uncalibrated images. To overcome the ambiguities that we mentioned in previous section, many approaches required the use of multiple images of the same subject with variable conditions (illumination, pose, expression, etc.) to constrain the number of possible solutions. \citep{KemelmacherICCV2011,SLiangECCV2016,RothCVPR2016}. However, gathering a collection of images from the same person is not always feasible. Other methods also explored 3D face reconstruction from a single image to overcome such limitation. In this case, different types of constraints have been presented to restrict the space of possible solutions. For instance, 3D facial templates were used in  \citep{KemelmacherPAMI2011,RothCVPR2015}, but they were highly dependant on the template's demographic. Most recent works based on photometric stereo methods \citep{XCaoCVPR2018,YLiCVMP2018,RotgerWSCG2019} combined their power to capture fine details with the capacity of 3DMMs to model global shape variations, obtaining impressive results.

The community has focused on developing methods that recover the 3D facial shape by fitting a 3DMM to the input image(s). The fitting process consists in estimating the model parameters that generate a 3D face as similar as possible to the one in the images. This is usually done by minimising the difference between the original image and the synthetic image obtained by rendering the 3DMM reconstruction, often guided by a set of corresponding 2D-3D points \citep{BoothPAMI2018,GecerCVPR2019,SariyanidiECCV2020}. Although this 3DMM-fitting approach has been widely explored with satisfactory results, leaning on the 3DMM as a hard constraint has its limitations. First, given the linear nature of 3DMMs, they are not able to model details but only global shape deformations, and hence the reconstructed 3D faces lack details. In addition, the accuracy of 3DMM-fitting methods may be limited by the type of information included in the model. For example, methods that are based on generating a synthetic image from the estimated 3D face usually use a 3DMM that also models the facial texture. When such a model is not available, either the facial texture is modelled separately or several sets of corresponding 2D-3D points are used to further guide the fitting process, such as facial landmarks, contour points \citep{QuBMVC2015,XZhuCVPR2015,PLiuMIPR2019} or image edges \citep{BasACCV2016}. Also, if the 3DMM does not model separately identity-related and expression-related deformations, during the fitting process, one type of deformation may be misinterpreted as the other, leading to sub-optimal reconstructions.

\subsection{Deep Learning}
\subsubsection{Training dataset}
Recent methods based on deep learning have shown to be highly successful in many other fields because of their ability to learn non-linear relationships of the data. However, they require large training datasets that are rarely available for the problem of 3D face reconstruction. To mitigate this problem, different training approaches have been proposed: 1) generation of synthetic 3D and 2D training datasets with the help of pre-trained 3D facial models \citep{XZhuCVPR2016,SelaICCV2017,YGuoPAMI2018,TranCVPR2018}, and 2) self-supervision by the minimisation of the difference between the input 2D image and the one rendered from the estimated 3D face, similarly to 3DMM-fitting methods \citep{TewariICCV2017,LTranCVPR2018,FWuCVPR2019,YChenTIP2020}. However, both approaches have their shortcomings. On the one hand, generating synthetic 2D and 3D data using 3DMMs does not overcome the lack of details since, as we have highlighted above, 3DMMs are not able to model subtle deformations. In addition, most 3DMMs are generated on insufficient data to capture the variability in the population, and also most models enforce constrains that are not necessarily realistic. On the other hand, self-supervision methods based on the synthetic generation of 2D renderings from a 3D facial reconstruction require accurate estimations of 3D facial textures, which is not trivia. 

\subsubsection{Architecture}
To alleviate these limitations of the training set, many different training schemes and network architectures have been proposed, such as iterative training \citep{XZhuPAMI2019,SanyalCVPR2019,BaiCVPR2020}, generative adversarial networks \citep{JLinCVPR2020,XTuToM2020,ZGaoCVPRW2020}, and training multiple networks each handling a specific task \citep{XFanToM2020,YChenTIP2020,XWangCVPR2020}. However, the approach that has been most exploited is the encoder-decoder architecture. This approach can be categorised into three groups depending on how the latent space of the encoder-decoder network is dealt with. The first category treats the encoder-decoder as a single network, not paying attention to the latent space that is learnt during training. For example, \citet{SelaICCV2017} and \citet{KoizumiECCV2020} trained an encoder-decoder network to estimate a depth map and a correspondence map from each pixel of the image to a vertex of a reference mesh. \citet{XZhuECCV2020} estimated a shape update that allowed them to recover finer details. The second category trains the encoder to extract meaningful image features that are then decoded into a 3D face. Among the works in this category, \citet{GenovaCVPR2018} and \citet{JSYoonCVPR2019} adopted pre-trained networks as encoders and trained the decoders to regress model parameters from the image features predicted by the encoders. The last category trains the encoder to learn descriptive features, such as features related to identity, expression, texture, etc. Most of the works that lie in this category, for example the work presented by \citet{LTranCVPR2019} and \citet{FLiuCVPR2018}, extract identity and non-identity features from the input images with a single encoder, and then decode each of them separately with two decoders. \citet{FLiuICCV2019} used the same idea of one encoder and two decoders, but they divided the training in two stages. The first one trains the architecture as an autoencoder using only 3D faces. The second one fine-tunes a pre-trained image feature extractor to predict the same latent features learned by the encoder trained in the first stage. In this second stage, the decoder trained in the first stage is kept fixed so that it reconstructs the corresponding 3D geometry. In contrast, \citet{XLiICASSP2020} argued that the extraction of identity and expression features are two separate tasks, and thus proposed two parallel encoder-decoders that separately estimate both features from local and global perspectives.

\subsection{Geometric Deep Learning}
A common drawback to all the deep learning approaches mentioned above is that classical deep learning operations can only deal with Euclidean data. However, 3D faces are usually represented as meshes, which is a non-Euclidean representation. For this reason, researchers have explored alternative representations such as depths maps \citep{SelaICCV2017,ZShuFG2020,KoizumiECCV2020}, volumetric representations \citep{JacksonICCV2017,HYiCVPR2019}, and parametric representations determined by 3DMMs \citep{XChaiICME2020,XTuToM2020,GenovaCVPR2018}. These representations of the 3D faces are often not accurate, and lose information with respect to the original 3D facial mesh, thus limiting the reconstruction potential.

Very recently, a new research field, called geometric deep learning \citep{BronsteinSPM2017_GeomDL}, has successfully extended classical deep learning operations to non-Euclidean data. In the case of 3D face reconstruction, adopting geometric deep learning methods allows for directly processing the 3D facial meshes, avoiding the need for intermediate Euclidean representations, and thus keeping their topological structure. In particular, the homologue to the widespread convolutional neural networks in this non-Euclidean framework are the graph convolutional networks (GCNs) \citep{BronsteinSPM2017_GeomDL,ZWuNNLS2021}, where the convolution operation is defined for non-Euclidean data. GCNs work with 3D facial meshes directly, and  have a small number of parameters, compared to CNNs, which makes them easier to train.

Several researchers have highlighted the advantages provided by GCNs for 3D facial analysis \citep{RanjanECCV2018,BouritsasICCV2019,ZJiangCVPR2019}. Their approach consisted in training a mesh-to-mesh autoencoder to learn a non-linear low-dimensional latent space that modelled facial shape deformations. Particularly, \citet{RanjanECCV2018} defined the convolution operator of their autoencoder GCN using spectral convolutional operators \citep{BrunaICLR2014} and filtered the meshes with kernels defined by the recursive Chebyshev polynomial \citep{DefferrardNIPS2016_GCN_spectral,HammondACHA2011_GCN_spectral}. With the same convolution operator, \citet{ZJiangCVPR2019} estimated two separate latent spaces: one that modelled identity-related shape variations, and another one that modelled expression-related deformations. Then they combined the reconstructions estimated from each of the latent spaces. Differently from the other two approaches, \citet{BouritsasICCV2019} highlighted that spectral convolutional operators are inherently isotropic. Therefore, they proposed a spiral convolution that is anisotropic by construction since it uses a pre-ordering of the mesh vertices following spiral trajectories, instead of taking patches formed by neighbouring vertices. This allowed them to have a one-to-one mapping between neighbours of a vertex and the parameters of the local filter.

Apart from 3D facial modelling, recently GCNs have also been incorporated into 3D face reconstruction systems to directly regress a 3D facial mesh. In particular, \citet{GHLeeCVPR2020}, \citet{ZGaoCVPRW2020}, \citet{JLinCVPR2020} and \citet{YZhouCVPR2019} designed a decoder using GCNs to directly recover a 3D facial mesh from image features extracted by an encoder. For example, the GCN decoder in \citep{GHLeeCVPR2020} translates uncertainty-aware features extracted by the encoder from the input image to a 3D face. \citet{ZGaoCVPRW2020} used two GCN decoders to regress the 3D face and per-vertex texture separately, and \citet{JLinCVPR2020} used GCNs to refine the per-vertex texture. Finally, the GCN decoder in \citep{YZhouCVPR2019} recovers a 3D face from features extracted using two separate encoders, one that inputs 3D faces, and another one that inputs 2D images. Thus, training such architecture end to-end, the encoders learn to extract the same latent features from both 2D images and 3D faces, whereas a single decoder learns to recover the 3D face.

The work we present in this paper is inspired by these latter approaches \citep{RanjanECCV2018,YZhouCVPR2019,FLiuICCV2019}, though it differs from them in several aspects: 1) \citep{RanjanECCV2018} focused on non-linear modelling of the 3D facial shape, whereas we integrate an autoencoder GCN within a 3D face reconstruction system; 2) \citep{YZhouCVPR2019} trained their 2D encoder from scratch, whereas we take advantage of powerful pre-trained networks to extract meaningful image features; and finally 3) \citep{FLiuICCV2019} used multi-layer perceptrons as the decoder, which are networks that are not formulated to deal with non-Euclidean data, whereas we exploit GCNs to directly work with 3D facial meshes. Furthermore, all of the above works address 3D face reconstruction for adults (or older children), but none of them focuses on babies as we do.
\section{BabyNet} \label{sec:3BabyNet}
\begin{figure}[h]
    \centering
    \includegraphics[width = \textwidth]{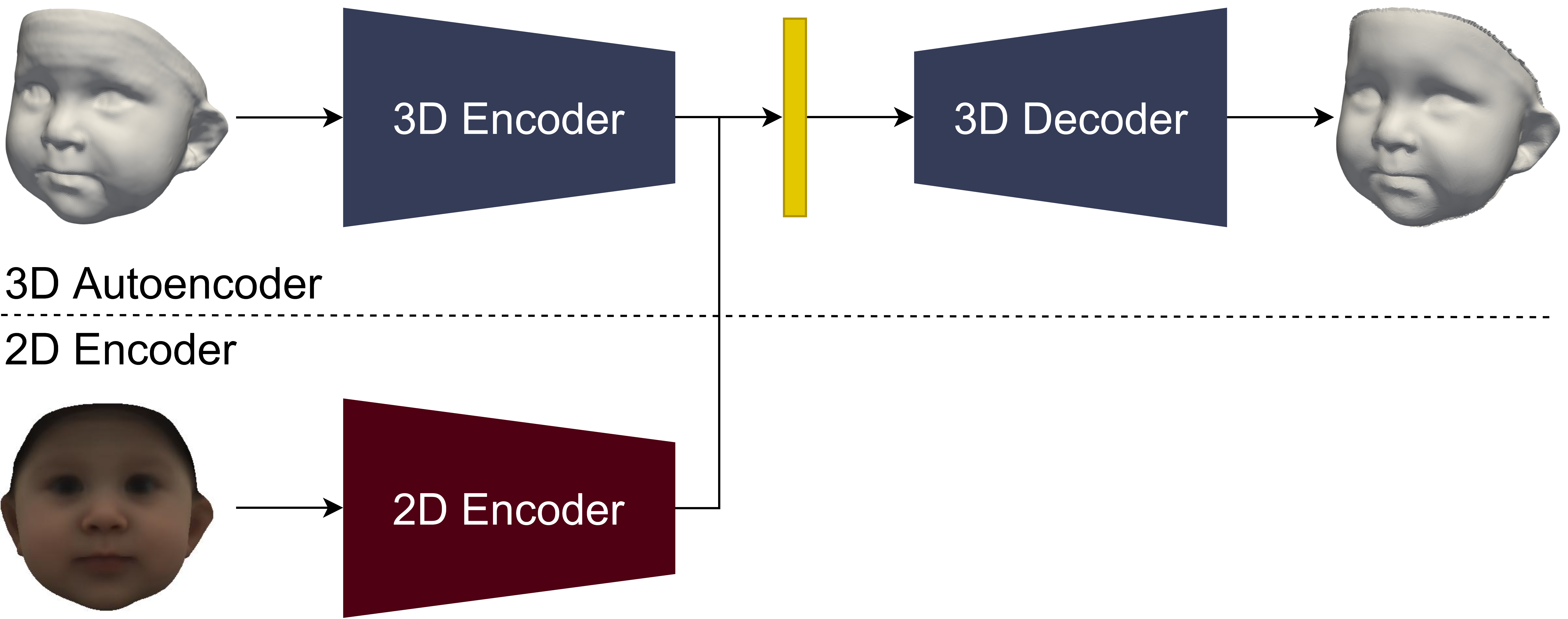}
    \caption{Global architecture of the proposed network. The two encoders (2D and 3D) estimate the same latent vector from corresponding 3D facial meshes and 2D pictures, and the 3D decoder recovers the 3D face from either 3D facial meshes or 2D images.}
    \label{fig:global_architecture}
\end{figure}

Our proposed architecture is designed following an encoder-decoder scheme with two differentiated parts: the \textit{3D autoencoder} and the \textit{2D encoder}, as can be seen in Figure \ref{fig:global_architecture}. In this way, the same features are extracted from corresponding 3D and 2D images, and both can be processes by the \textit{3D decoded} to reconstruct the 3D face. In this section, we describe each of the components separately and provide implementation details.

\subsection{3D autoencoder} \label{subsec:3Dautoencoder_arch}
To adequately handle facial surfaces represented as 3D meshes, we design the 3D autoencoder using GCNs for both the encoder and the decoder. As explained in Section \ref{sec:2RelatedWork}, this type of network redefines the classical operations found in convolutional neural networks to handle non-Euclidean data, such as 3D facial meshes, by reformulating convolution and sampling operations as explained next.

\textbf{Mesh convolution.} We adopted the mesh convolution method proposed in \cite{BrunaICLR2014} based on the spectral decomposition of the graph Laplacian. Let $\mathcal{V}\subset \mathbb{R}^3$ be the set of $N$ vertices in a 3D facial mesh, and $\mathbf{A} \in \{0,1\}^{N\times N}$ its adjacency matrix. The normalised Laplacian of this mesh is defined as
\begin{equation*}
    \mathbf{L} = \mathbf{I}_N - \mathbf{D}^{-\frac{1}{2}}\mathbf{A}\mathbf{D}^{-\frac{1}{2}},
\end{equation*}
where $\mathbf{I}_N$ is the $N\times N$ identity matrix, and $\mathbf{D}$ is the diagonal matrix containing the degree of each vertex, i.e., $\mathbf{D}_{ii} = \sum_{j=1}^N \mathbf{A}_{ij}$. Since $\mathbf{L}$ is real symmetric by construction, the spectral theorem states that it can be diagonalised, i.e., there exist a diagonal matrix $\mathbf{\Lambda} \in \mathbb{R}^{N \times N}$ and an orthonormal matrix $\mathbf{U} \in \mathbb{R}^{N\times N}$ such that $\mathbf{L = U\Lambda U^\text{T}}$. In the orthonormal basis defined by the eigenvectors of $\mathbf{L}$, i.e, the columns of $\mathbf{U}$, the graph convolution operation $*_\mathcal{G}$ of a mesh $\mathbf{x}\in\mathbb{R}^N$ by a filter $\mathbf{g}$ is defined as
\begin{equation*}
\mathbf{x}*_\mathcal{G}\mathbf{g} = \mathbf{U}\left( \mathbf{U}^\text{T}\mathbf{x} \odot \mathbf{U}^\text{T}\mathbf{g} \right),
\end{equation*}
where $\odot$ is the Hadamard product. Thus, denoting $\mathbf{g_w} := \text{diag}(\mathbf{U}^\text{T}\mathbf{g}) \in \mathbb{R}^{N\times N}$, the graph convolution is simplified as a straight-forward product in the spectral domain, which can be translated back to the graph domain by simple matrix product:
\begin{equation}\label{eq:graphconv}
\mathbf{x}*_\mathcal{G}\mathbf{g}_\mathbf{w} = \mathbf{U}\left( \mathbf{g}_\mathbf{w}\mathbf{U}^\text{T}\mathbf{x}\right).
\end{equation}
In this context, $\mathbf{g_w}$ can be interpreted as the coefficients of a filter in the spectral domain.

Although the choice of $\mathbf{g_w}$ is not unique, the filter proposed by \citet{DefferrardNIPS2016_GCN_spectral} has been widely used, which uses a truncated expansion of the Chebyshev polynomials $\{T_k(\cdot)\}_{k=0}^K$ to parametrise  $\mathbf{g_w}$ as:
\begin{equation*}
    \mathbf{g_w} = \sum_{k=0}^K w_k T_k(\widetilde{\mathbf{\Lambda}}),
\end{equation*}
where the coefficients $w_k$ are the weights that are learnt by the network in the training phase, and $\widetilde{\mathbf{\Lambda}} = 2\mathbf{\Lambda}/\lambda_\text{max} - \mathbf{I}_N$, with $\lambda_\text{max}$ the highest eigenvalue of $\mathbf{L}$. Therefore, denoting $\widetilde{\mathbf{L}} = \mathbf{U}\widetilde{\mathbf{\Lambda}}\mathbf{U}^\text{T}$, the spectral graph convolution from eq. \eqref{eq:graphconv} becomes \footnote{It can be demonstrated that $T_k(\widetilde{\mathbf{L}}) = \mathbf{U}T_k(\widetilde{\mathbf{\Lambda}})\mathbf{U}^\text{T}$, and that $\widetilde{\mathbf{L}} = 2\mathbf{L}/\lambda_\text{max} - \mathbf{I}_N$.}
\begin{equation}\label{eq:graphconv_cheb}
\mathbf{x}*_\mathcal{G}\mathbf{g}_\mathbf{w} = \sum_{k=0}^K w_k T_k(\widetilde{\mathbf{L}})\mathbf{x}.
\end{equation}

This choice of the filter greatly simplifies the computations since the eigendecomposition of the graph Laplacian is avoided, and it is replaced by multiplications of sparse matrices.

\textbf{Mesh down-sampling and up-sampling.} Similar to the work presented in \citep{RanjanECCV2018}, we use quadric matrices \citep{GarlandSIGGRAPH1997} to minimise the surface error during down-sampling of our meshes. In addition, we compute the down-sampling $\mathbf{Q}_d$ and up-sampling $\mathbf{Q}_u$ matrices as a pre-processing step before training as explained next.

Let us assume that we are down-sampling a mesh from a set $\mathcal{V}_1$ of $N_1$ vertices to a set of $\mathcal{V}_2$ of $N_2$ vertices, where $N_1 > N_2$. Let $\mathbf{v}_{i_1} \in \mathcal{V}_1$ and $\mathbf{v}_{i_2} \in \mathcal{V}_2$. Then, we can define a down-sampling matrix $\mathbf{Q}_d \in \{0,1\}^{N_2 \times N_1}$ for $i_1 = 1,\cdots,N_1$ and $i_2 = 1, \cdots, N_2$ as
\begin{equation*}
\mathbf{Q}_d (i_2, i_1) = \left\{
\begin{array}{ll}
    1 \text{ if vertex } \mathbf{v}_{i_1} \text{ is kept as $\mathbf{v}_{i_2} \in \mathcal{V}_2$,}\\
    0 \text{ otherwise.}
\end{array}\right.
\end{equation*}
Therefore, the down-sampled shape $\mathbf{s}_2 \in \mathbb{R}^{N_2 \times 3}$ is computed from the original shape $\mathbf{s}_1 \in \mathbb{R}^{N_1 \times 3}$ using sparse matrix multiplication: $\mathbf{s}_2 = \mathbf{Q}_d \mathbf{s}_1$.

Then, the up-sampling matrix $\mathbf{Q}_u \in [0,1]^{N_1 \times N_2}$ from $\mathcal{V}_2$ to $\mathcal{V}_1$ is defined as $\mathbf{Q}_u(i_1,i_2) = 1$ for the vertices $\mathbf{v}_{i_1} \in \mathcal{V}_1$ that were kept in the down-sampling process, i.e, $\mathbf{Q}_d(i_2,i_1) = 1$. The vertices $\mathbf{v}_{i_1} \in \mathcal{V}_1$ that were discarded are recovered by projecting them into the closest triangle in the down-sampled mesh $(\mathbf{v}_{i_2},\mathbf{v}_{j_2},\mathbf{v}_{k_2}) \subset \mathcal{V}_2$, and expressing them in barycentric coordinates, $\widetilde{\mathbf{v}}_{i_1} = r_i \mathbf{v}_{i_2} + r_j \mathbf{v}_{j_2} + r_k \mathbf{v}_{k_2}$, where $\widetilde{\mathbf{v}}_{i_1}$ denotes the projection of $\mathbf{v}_{i_1}$. Then, the up-sampling matrix is defined for these vertices as $\mathbf{Q}_u(i_1,i_2) = r_i$, $\mathbf{Q}_u(i_1,j_2) = r_j$, and $\mathbf{Q}_u(i_1,k_2) = r_k$. To sum up, the up-sampling matrix is defined as
\begin{equation*}
\mathbf{Q}_u (i_1, i_2) = \left\{
\begin{array}{ll}
    1 &\text{ if vertex } \mathbf{v}_{i_1} \text{ was kept in the down-sampling,}\\
    r_i &\text{ if vertex } \mathbf{v}_{i_1} \text{ was not kept, and $\widetilde{\mathbf{v}}_{i_1} = r_i \mathbf{v}_{i_2} + r_j \mathbf{v}_{j_2} + r_k \mathbf{v}_{k_2}$,}\\
    0 &\text{ otherwise.}
\end{array}\right.
\end{equation*}
Therefore, the original shape $\mathbf{s}_1 \in \mathbb{R}^{N_1 \times 3}$ can be recovered from the down-sampled shape $\mathbf{s}_2 \in \mathbb{R}^{N_2 \times 3}$ as $\mathbf{s}_1 = \mathbf{Q}_u \mathbf{s}_2$.

\begin{figure}[h]
    \centering
    \includegraphics[width = \textwidth]{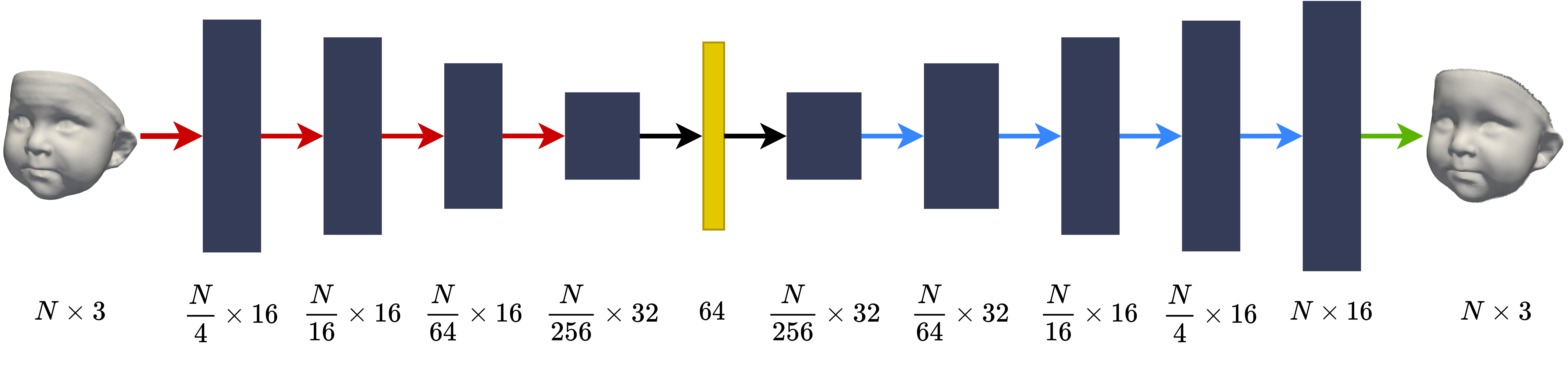}
    \caption{Architecture of the proposed \textit{3D autoencoder}. Red and blue arrows indicate graph convolutional layers with down-sampling and up-sampling, respectively. Black arrows indicate fully connected layers, and the green arrow indicates a graph convolutional layer without up-sampling. The yellow block represents the latent space.}
    \label{fig:3Dautoencoder_arch}
\end{figure}

Following \cite{RanjanECCV2018} and the mesh operations described above, we designed a \textit{3D autoencoder} with five layers in the encoder, and six layers in the decoder, as shown in Figure \ref{fig:3Dautoencoder_arch}. We used a sampling factor of 4 for all the layers with mesh sampling, similarly to \citep{RanjanECCV2018}. However, unlike \citep{RanjanECCV2018}, we adopted a graph convolutional filter with the expansion of the Chebyshev polynomials truncated
at $K = 9$ (see eq. \eqref{eq:graphconv_cheb}), which improved our results (see Section \ref{sec:5SystemEvaluation}).

\subsection{2D encoder}\label{subsec:2Dencoder_arch}
The \textit{2D encoder} is initialised using transfer learning to exploit the discriminative power of pre-trained networks. As the base encoder, we adopted a recently published network, the ArcFace \citep{DengCVPR2019_arcface}, which trained a ResNet-50 architecture for a facial recognition task using a novel loss function that provided lower recognition errors than other state-of-the-art methods.

\begin{figure}[ht]
    \centering
    \includegraphics[width = \textwidth]{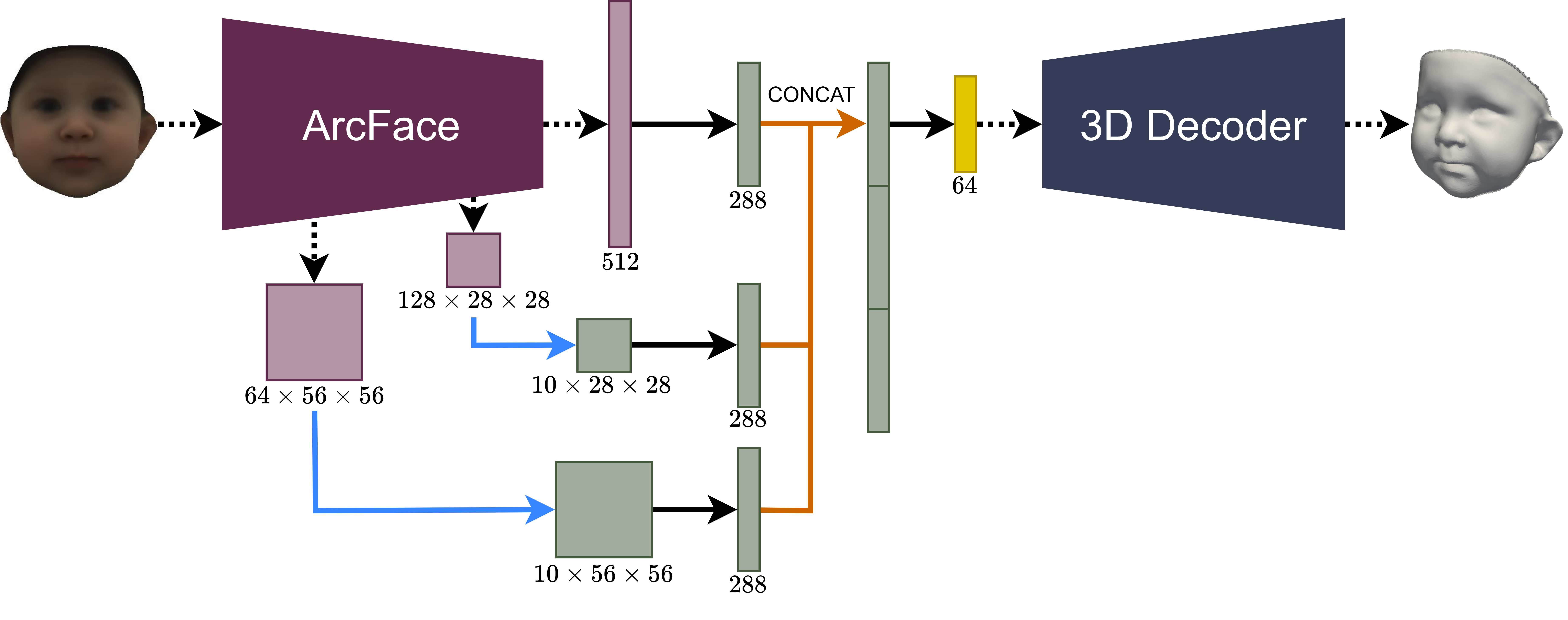}
    \caption{Architecture of the proposed \textit{2D encoder}. Dotted grey arrows indicate ``input'' or ``output'', blue arrows indicate 1D convolutional layers, black arrows indicate fully connected layers, and orange arrows indicate concatenation.}
    \label{fig:2Dencoder_arch}
\end{figure}

The ArcFace estimates from the input image a feature vector of dimension 512, but our \textit{3D decoder} inputs feature vectors of dimension $64$. Therefore, we have to learn a mapping from the ArcFace's latent space to the \textit{3D autoencoder} latent space, so that we can use the \textit{3D decoder} to recover a 3D face. To do so, we used both the 512-feature vector from the ArcFace and the feature maps estimated by intermediate layers. This approach allowed us to capture finer details to estimate a 64-feature latent vector used for reconstructing the 3D facial geometry by the \textit{3D decoder}. More specifically, this global-local approach is designed as follows: 1) the dimensionality of the feature vector estimated by the output layer is reduced by a fully connected layer; 2) the feature maps extracted by the 7th and the 15th layers, which are the last feature maps of size $64\times56\times56$ and $128\times28\times28$ estimated by ArcFace, respectively, are mapped to a feature vector by first reducing the number of channels using 1D convolutional layers and then using fully connected layers; finally, 3) the three feature vectors (a global one estimated from the 512-feature vector, and two local ones estimated from the feature maps) are concatenated and passed through another fully connected layer that estimates the final 64-feature vector, which is decoded into a 3D face by the \textit{3D decoder}. This architecture is illustrated in Figure \ref{fig:2Dencoder_arch}.

\section{Training procedure} \label{sec:4Training}
In this section, we describe how the training data was generated, and then, we specify the details of the training.

\subsection{Training data} \label{subsec:TrainingData}
As explained in Section \ref{sec:2RelatedWork}, the main state-of-the-art strategies to address the limited 3D facial image datasets in adults can be divided into methods that generate synthetic training sets with the help of 3DMMs, and methods that use self-supervision-based techniques that create synthetic images from the 3D facial reconstructions. Although both approaches have limitations, we believe that self-supervision makes the reconstruction process much more complex, since it incorporates also the texture into the 3D face reconstruction problem. Consequently, since the network has to learn two tasks, it is harder to train and to obtain satisfactory results. For this reason, we follow the first approach and train our network with synthetically generated data using the BabyFM \citep{MoralesFG2020,MoralesPAMI2022}.

Our training dataset consists of $20,000$ data pairs of 3D faces and 2D pictures sampled from the BabyFM. These 3D faces were sampled from the BabyFM shape model, and, for each of these synthetically generated face shapes, a texture was sampled using the BabyFM texture model. Then, the textured 3D faces are rendered to obtain corresponding frontal 2D pictures, following a perspective projection. The validation set of 100 data samples is created following the same procedure as the training set. 

\subsection{Training details}
\textit{\textbf{3D autoencoder.}} Figure \ref{fig:3Dautoencoder_arch} shows the main architecture details of the \textit{3D autoencoder}. We use ReLU activations to all the spectral convolutional layers with mesh sampling (blue and red arrows in Figure \ref{fig:3Dautoencoder_arch}) and the fully connected layers (black arrows in Figure \ref{fig:3Dautoencoder_arch}). The \textit{3D autoencoder} was trained during $300$ epochs with a stochastic gradient descent (SGD) optimiser. We set the momentum to $0.9$, the weight decay to $0.0005$, and the learning rate to $0.008$. The learning rate was adjusted after each epoch with a rate decay of $0.98$. We use a batch size of 16 samples, and the $L_1$ loss function between the input and the output 3D faces, as it produced more accurate results. Among the $300$ epochs, the model we selected is the one performing the best on the validation set.


\textit{\textbf{2D encoder.}} Figure \ref{fig:2Dencoder_arch} shows the architecture of the \textit{2D encoder}. All the fully connected layers of the \textit{2D encoder} (black arrows in Figure \ref{fig:2Dencoder_arch}) were preceded by a dropout layer with dropout probability of $0.25$, and followed by a ReLU activation function. The \textit{2D encoder} was also trained during 300 epochs with a SGD optimiser. The momentum was also set to $0.9$ and the learning rate was set to $0.01$. No weight decay was used, nor learning rate decay. The batch size was also set to 16, and $L_1$ loss function was used to compare the latent space representations estimated by the \textit{2D encoder} and the \textit{3D encoder}. Again, the selected model was the one performing best on the validation set among the 300 epochs.


\section{System evaluation} \label{sec:5SystemEvaluation}
In this section, we evaluate the major design choices of both the \textit{3D autoencoder} (see Section \ref{subsec:3Dautoencoder_arch}) and the \textit{2D encoder} (see Section \ref{subsec:2Dencoder_arch}). 

\subsection{3D autoencoder}
To design our \textit{3D autoencoder}, we made two major changes compared to the original work presented in \citep{RanjanECCV2018}: 1) we increased the latent space from 8 to 64 features to capture more information needed to reconstruct the 3D faces accurately; and 2) we approximated our filters using the first 9 Chebyshev polynomials instead of 6 to allow us to use a higher order filter and a larger neighbourhood to process each vertex, thus capturing the facial details.

We fine-tuned our \textit{3D autoencoder} using down-sampled versions of the training and the validation dataset, decreasing the number of vertices from $31,027$ to $5,273$, which allowed us to decrease the computational time to a fifth. The results shown in Tables \ref{tab:3Dautoencoder_latentSize} and \ref{tab:3Dautoencoder_K} are obtained with the down-sampled datasets. Although the final model is trained with the training set with the original number of vertices, and the results shown in Section \ref{sec:6CompWithSOTA} are obtained with this final model.

Table \ref{tab:3Dautoencoder_latentSize} shows the mean Euclidean error between the input and the output facial meshes for different sizes of the latent space. Increasing the size of the latent space to 64 shows a significant improvement with respect to size of 8 proposed by \citep{RanjanECCV2018} (with $p$-value $< 0.001$). This suggests that a latent size of 8 is not sufficient to capture the facial details of the input mesh. However, a higher-dimensional latent feature space, as we show for a latent space size of 128, does not provide any accuracy improvement, whereas it increases the computational complexity of the network.
\begin{table}[h]
\centering
\caption{Mean Euclidean errors (MEE) in mm over the validation set obtained with the \textit{3D autoencoder} for different latent sizes.} \label{tab:3Dautoencoder_latentSize}
\begin{tabularx}{\textwidth}{l|Y|Y|Y|Y|Y}
Latent sz.          & 8 & 16 & 32 & 64 & 128  \\ \hline
MEE & $1.26 \pm 0.24$ & $0.94 \pm 0.13$ & $0.57\pm 0.06$ & $0.31\pm 0.05$ & $0.31\pm 0.05$
\end{tabularx}
\end{table}

For a fixed latent space of size 64, we evaluated the effect of the number of Chebyshev polynomials that define the graph convolutional filter. As it can be seen in Table \ref{tab:3Dautoencoder_K}, using higher order polynomials slightly improved the results, as a larger neighbourhood is used to filter each vertex. However, it also increases the computational complexity. We believe that truncating the Chebyshev polynomials expansion at $K = 9$ (see eq. \eqref{eq:graphconv_cheb}) is a good compromise between the computational complexity and the improvement of the accuracy of the estimated facial 
\begin{table}[h]
\centering
\caption{Mean Euclidean errors (MEE) in mm over the validation set obtained with the \textit{3D autoencoder} for different values for the truncation of the Chebyshev polynomials ($K$ in eq. \eqref{eq:graphconv_cheb}) defining the graph convolutional filter.} \label{tab:3Dautoencoder_K}
\begin{tabularx}{\textwidth}{l|Y|Y|Y|Y|Y}
$K$ & 6 & 7 & 8 & 9 & 10   \\ \hline
MEE & $0.34 \pm 0.06$ & $0.34 \pm 0.06$ & $0.32\pm 0.05$ & $0.31\pm 0.05$ & $0.30\pm0.05$
\end{tabularx}
\end{table}
meshes ($p$-value $< 0.001$ against $K=6$). Furthermore, the decrease in the errors obtained with $K = 10$ is not significant ($p$-value$=0.043$ against $K=7$, and $p$-value$=1$ against $K=8$).


\subsection{2D encoder}
\begin{figure}[ht]
    \begin{subfigure}[t]{0.3\textwidth}
        \centering
        \includegraphics[width=0.6\textwidth]{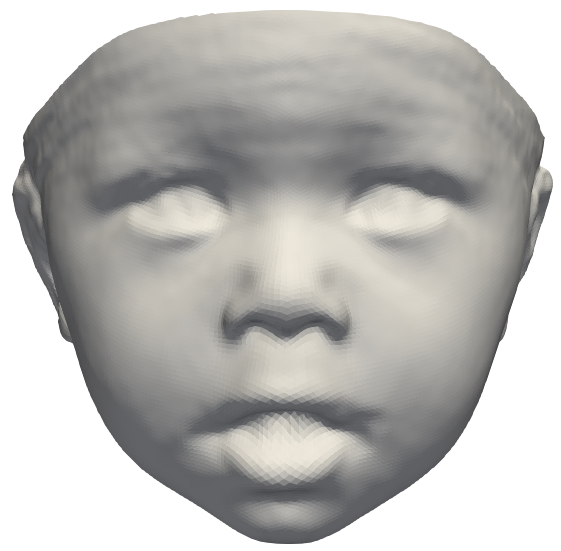}
    \end{subfigure}
    \hfill
    \begin{subfigure}[t]{0.3\textwidth}
        \centering
        \includegraphics[width=0.6\textwidth]{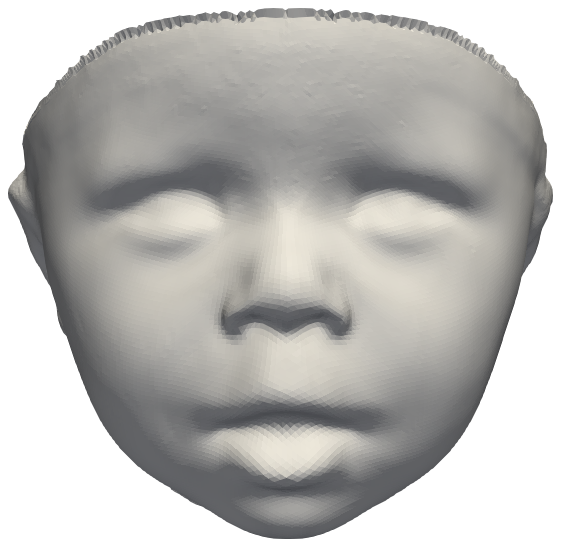}
    \end{subfigure}
    \hfill
    \begin{subfigure}[t]{0.3\textwidth}
        \centering
        \includegraphics[width=0.6\textwidth]{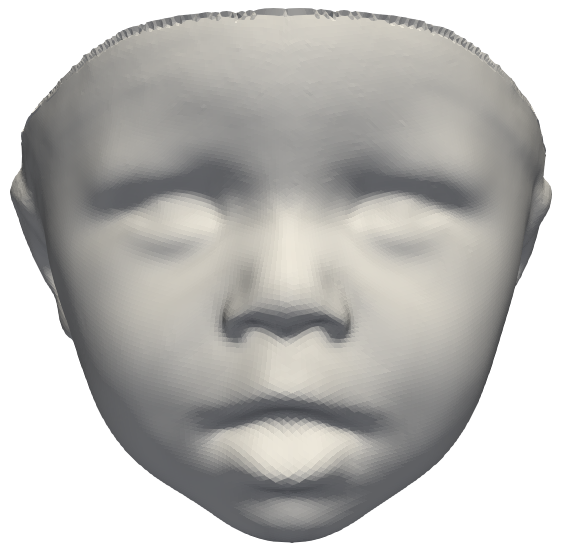}
    \end{subfigure}
    \hfill
    \begin{subfigure}[t]{0.3\textwidth}
        \centering
        \includegraphics[width=0.6\textwidth]{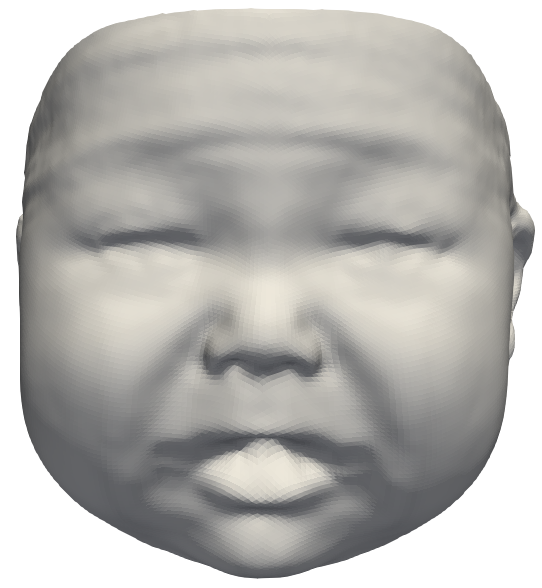}
    \end{subfigure}
    \hfill
    \begin{subfigure}[t]{0.3\textwidth}
        \centering
        \includegraphics[width=0.6\textwidth]{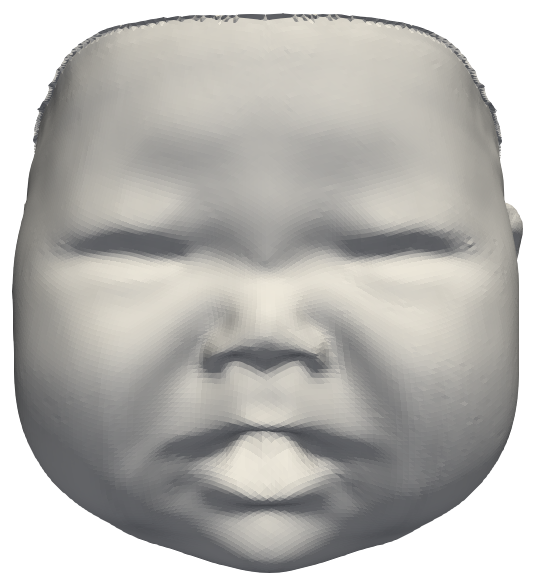}
    \end{subfigure}
    \hfill
    \begin{subfigure}[t]{0.3\textwidth}
        \centering
        \includegraphics[width=0.6\textwidth]{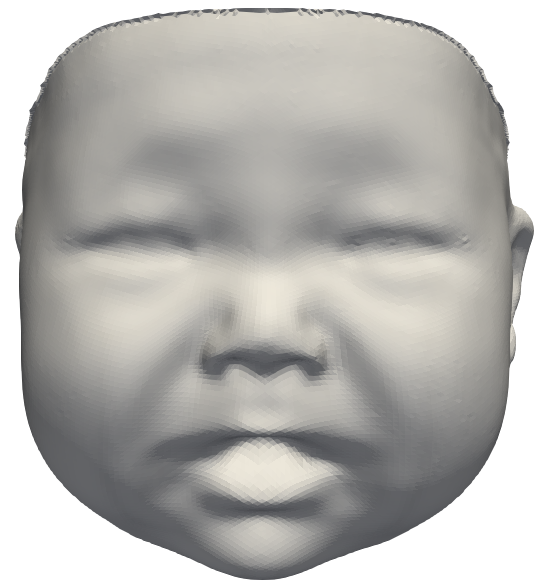}
    \end{subfigure}
    \hfill
    \begin{subfigure}[t]{0.3\textwidth}
        \centering
        \includegraphics[width=0.6\textwidth]{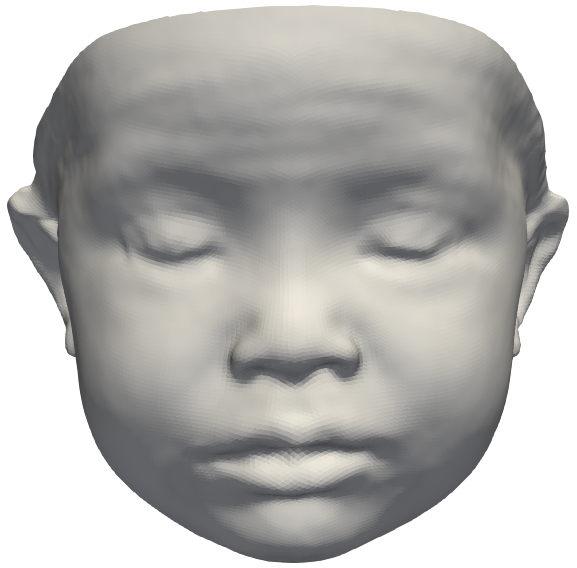}
        \caption{Ground truth 3D face.} \label{fig:2Dencoder_intermediateLayers_gt}
    \end{subfigure}
    \hfill
    \begin{subfigure}[t]{0.3\textwidth}
        \centering
        \includegraphics[width=0.6\textwidth]{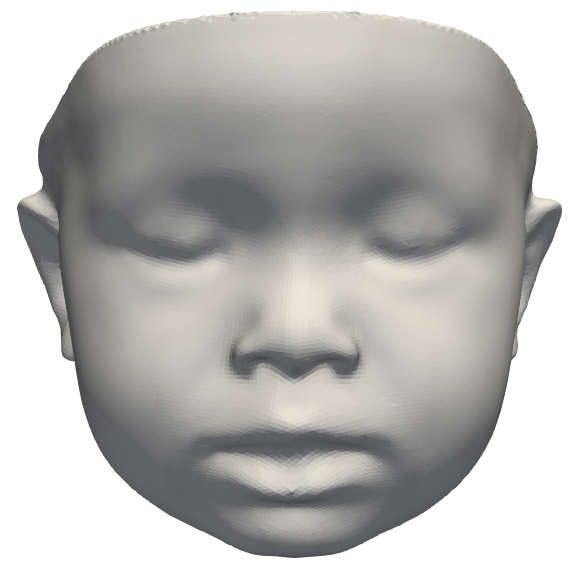}
        \caption{With features from output layer.} \label{fig:2Dencoder_intermediateLayers_P14}
    \end{subfigure}
    \hfill
    \begin{subfigure}[t]{0.3\textwidth}
        \centering
        \includegraphics[width=0.6\textwidth]{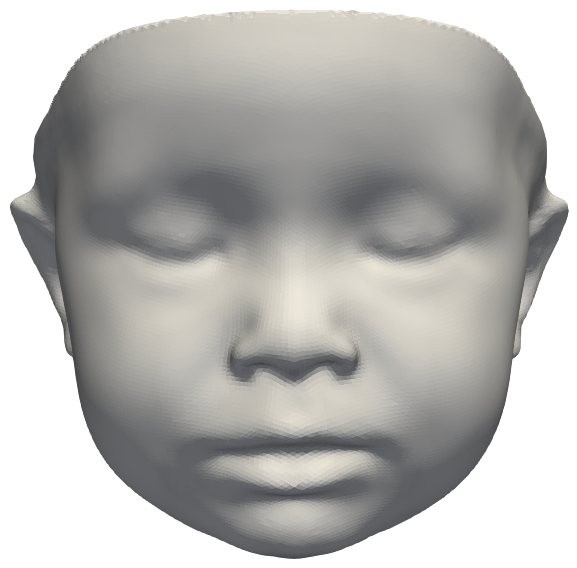}
        \caption{With features from output and intermediate layers.} \label{fig:2Dencoder_intermediateLayers_P29}
    \end{subfigure}
    \hfill
    \caption{Examples of 3D facial reconstructions from the validation set. (b) Only features from the output layer of the ArcFace were extracted. (c) Features from the output layer and from the 7th and the 15th layers were extracted.}
    \label{fig:2Dencoder_intermediateLayers}
\end{figure}

Our first approach was to use fully connected layers to map the image features extracted by the output layer of the ArcFace \citep{DengCVPR2019_arcface} to the latent space of the \textit{3D autoencoder}. We show the resulting mean Euclidean errors for different numbers of fully connected layers in Table \ref{tab:2Dencoder_FC}. As it can be seen, the use of 2 fully connected layers improves the results obtained with 1 fully connected layer with a $p$-value $< 0.001$. Even though, using 3 layers slightly reduces the standard deviation, the improvement does not compensate for the increase in computational complexity with respect to using 2 layers. Even so, the resulting 3D facial reconstructions obtained with 2 fully connected layers were unrealistically smooth and lacked fine details, as shown in Figure \ref{fig:2Dencoder_intermediateLayers_P14}.

\begin{table}[h]
\centering
\caption{Mean Euclidean errors (MEE) in mm over the validation set obtained with the \textit{2D encoder}+\textit{3D decoder} for different number of fully connected layers to map the image features extracted by the output layer of the ArcFace to the latent feature of the \textit{3D autoencoder}.} \label{tab:2Dencoder_FC}
\begin{tabular}{l|c|c|c}
No. fully connected layers & 1    & 2    & 3    \\ \hline
MEE             & $3.22 \pm 1.40$ & $1.82 \pm 0.54$ & $1.82 \pm 0.48$
\end{tabular}
\end{table}

Table \ref{tab:2Dencoder_intermediateLayers_val} shows how incorporating intermediate feature maps in our \textit{2D encoder} substantially improves the results compared to using only features predicted by the output layer. Note that all the errors shown in Table \ref{tab:2Dencoder_intermediateLayers_val} are lower than the ones shown in Table \ref{tab:2Dencoder_FC}. This is also illustrated in Figure \ref{fig:2Dencoder_intermediateLayers_P29}, where we qualitatively show that this approach provides more accurate and realistic reconstructions.

\begin{table}[h!]
\centering
\caption{Mean Euclidean errors (MEE) in mm over the validation set obtained with the \textit{2D encoder}+\textit{3D decoder} with feature maps extracted from different number of intermediate layers, in addition to the output layer.} \label{tab:2Dencoder_intermediateLayers_val}
\begin{tabular}{l|c}
Features from intermediate layers & MEE \\ \hline
1 (7th) & $1.13 \pm 0.30$ \\ \hline
1 (15th) & $0.98 \pm 0.27$ \\ \hline
1 (45th) & $1.07 \pm 0.29$ \\ \hline
2 (7th and 15th) & $1.01 \pm 0.28$ \\ \hline
2 (7th and 45th) & $0.95 \pm 0.26$ \\ \hline
2 (15th and 45th) & $0.88 \pm 0.24$ \\ \hline
3 (7th, 15th, and 45th) & $0.88\pm 0.26$
\end{tabular}
\end{table}

To determine which intermediate feature maps provide better results, we evaluated the combination of features at three different scales: $64\times56\times56$ (extracted from the 7th layer), $128\times28\times28$ (extracted from the 15th layer), and $256\times 14 \times 14$ (extracted from the 45th layer). The last feature map estimated by the ArcFace, which is of size $512\times 7 \times 7$, was not used since it encodes too high-level features and thus the information captured is not as detailed as needed. As can be seen in Table \ref{tab:2Dencoder_intermediateLayers_val}, combining several scales of feature maps provides better results than using a single scale. Also, it seems that using higher level features (extracted by deeper layers), which encode more global information, produce more accurate results. However, as mentioned in Section \ref{subsec:TrainingData}, the validation set is fully synthetic, and thus it is not as detailed as a real dataset. This introduces a bias in the experiment, since higher level features are more useful to reconstruct smoother geometries, whereas lower level features that encode finer details only contribute with noise to the reconstruction. On the contrary, such lower level features are advantageous when targeting real datasets, which are more detailed. We show this in Table \ref{tab:2Dencoder_intermediateLayers_test}, where the same architectures as in Table \ref{tab:2Dencoder_intermediateLayers_val} are evaluated over a dataset of real 3D facial images of babies (see Section \ref{subsec:TestSet}). In Table \ref{tab:2Dencoder_intermediateLayers_test}, we can see that the best accuracy is obtained when features extracted by shallower layers, i.e., the 7th and the 15th layers, are combined. Adding also the feature map from the 45th layer, the improvement of the accuracy is minimal but the computational complexity increases considerably.

\begin{table}[h!]
\centering
\caption{Mean Euclidean errors (MEE) over the test set obtained with the \textit{2D encoder}+\textit{3D decoder} with feature maps extracted from different number of intermediate layers, in addition to the output layer.} \label{tab:2Dencoder_intermediateLayers_test}
\begin{tabular}{l|c}
Features from intermediate layers & MEE \\ \hline
1 (7th) & $3.10 \pm 1.12$ \\ \hline
1 (15th) & $3.13 \pm 1.09$ \\ \hline
1 (45th) & $3.19 \pm 1.09$ \\ \hline
2 (7th and 15th) & $3.04 \pm 1.12$ \\ \hline
2 (7th and 45th) & $3.06 \pm 1.14$ \\ \hline
2 (15th and 45th) & $3.07 \pm 1.11$ \\ \hline
3 (7th, 15th, and 45th) & $3.01\pm 1.10$
\end{tabular}
\end{table}


\section{Comparison with the state-of-the-art} \label{sec:6CompWithSOTA}

In this section, we compare the reconstruction accuracy of the BabyNet to other state-of-the-art methods using an independent test dataset.

\subsection{Test dataset} \label{subsec:TestSet}
\begin{figure}[h]
    \begin{subfigure}[t]{0.3\textwidth}
        \centering
        \includegraphics[width=0.8\textwidth]{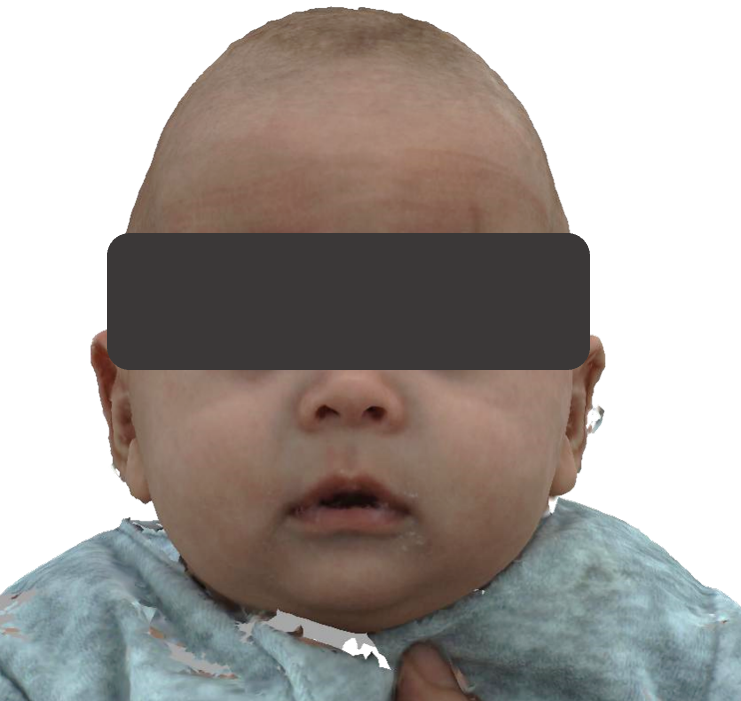}
    \end{subfigure}
    \hfill
    \begin{subfigure}[t]{0.3\textwidth}
        \centering
        \includegraphics[width=0.8\textwidth]{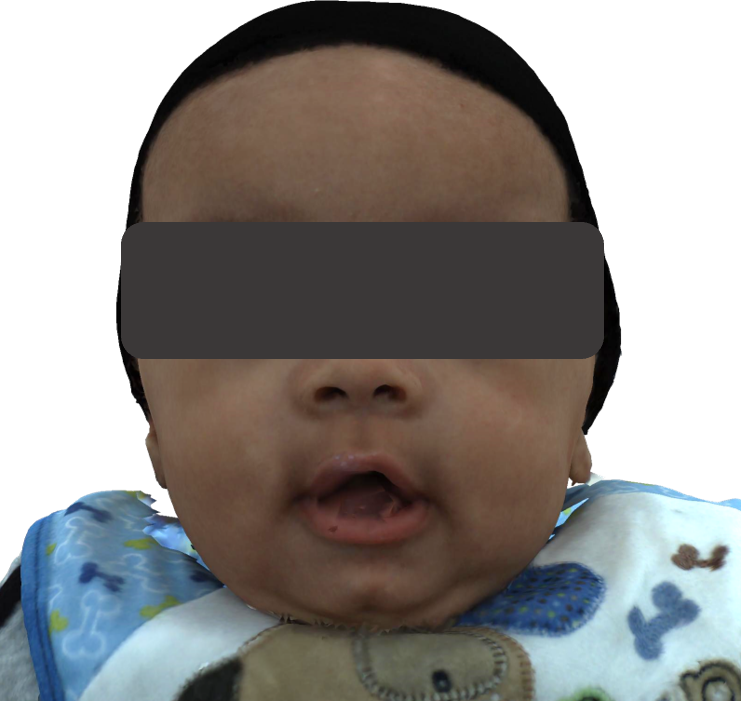}
    \end{subfigure}
    \hfill
    \begin{subfigure}[t]{0.3\textwidth}
        \centering
        \includegraphics[width=0.8\textwidth]{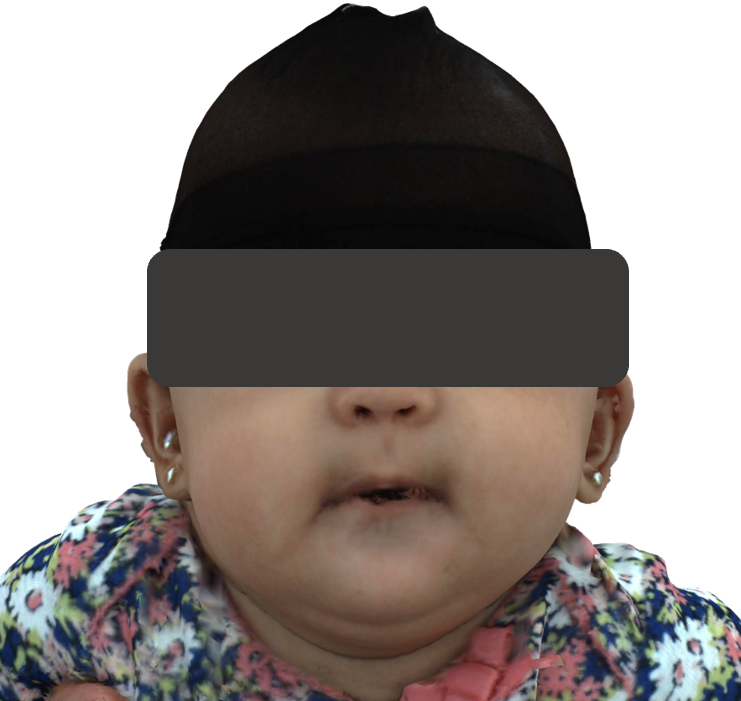}
    \end{subfigure}
    \hfill
    \begin{subfigure}[t]{0.3\textwidth}
        \centering
        \includegraphics[width=0.8\textwidth]{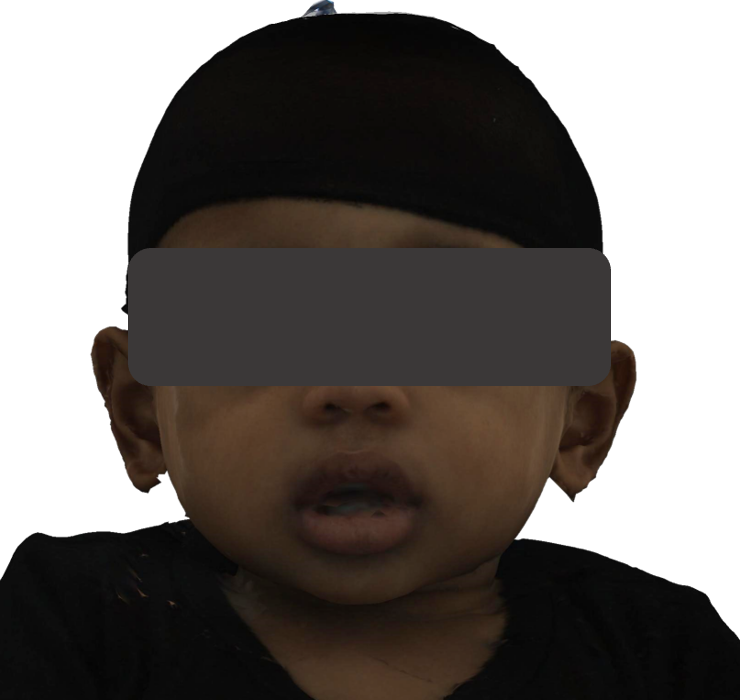}
    \end{subfigure}
    \hfill
    \begin{subfigure}[t]{0.3\textwidth}
        \centering
        \includegraphics[width=0.8\textwidth]{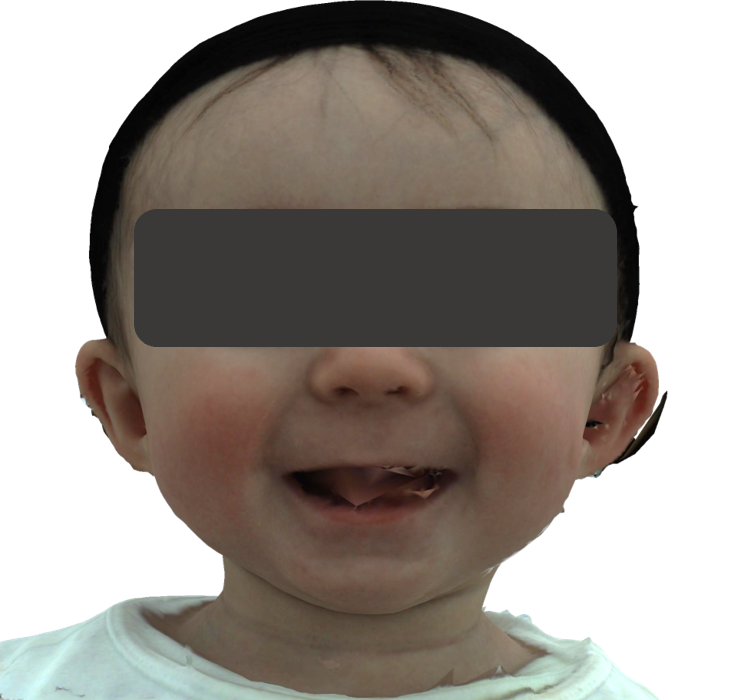}
    \end{subfigure}
    \hfill
    \begin{subfigure}[t]{0.3\textwidth}
        \centering
        \includegraphics[width=0.8\textwidth]{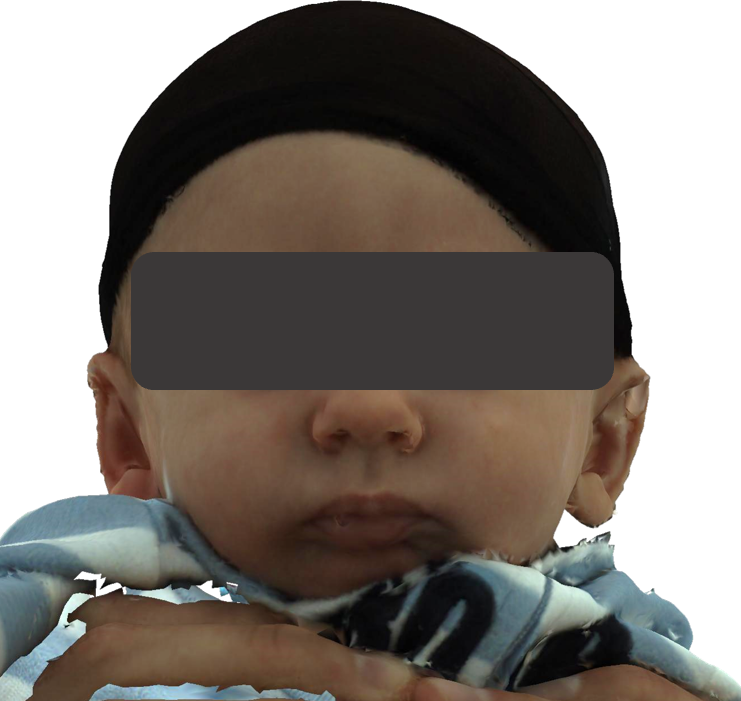}
    \end{subfigure}
    \hfill
    \caption{Examples of 2D images rendered from the textured 3D facial scans in the test set.}
    \label{fig:examples_2Dimages}
\end{figure}

The test set used to carry out the aforementioned comparison contains 70 3D scans provided by the Children's National Hospital in Washington D.C., which we used to render corresponding 2D images with highly visual resemblance to real pictures, as shown in Figure \ref{fig:examples_2Dimages}. This dataset contains babies of ages between 1 and 34 months old, it is roughly gender balanced, and several ethnicities are included.

\subsection{Experimental setup} \label{subsec:ExpSetup}
We compare the performance of the BabyNet with two recently published deep learning 3D face reconstruction methods that are publicly available: 3DDFA\_V2 \citep{JGuoECCV2020}\footnote{\url{ https://github.com/cleardusk/3DDFA_V2}} and DFNRMVS \citep{BaiCVPR2020}\footnote{\url{https://github.com/zqbai-jeremy/DFNRMVS}}. Since these methods were trained with an adult dataset, with this comparison, we aim to show the need for a 3D face reconstruction system specific for babies. In addition, we also compare the performance of the BabyNet with a publicly available classical 3DMM-fitting method proposed by \citep{BasACCV2016}\footnote{\url{https://github.com/waps101/3DMM_edges}}, but using the BabyFM \citep{MoralesFG2020,MoralesPAMI2022}.

To compute the reconstruction errors, we follow a similar approach to the one proposed in \citep{MoralesFG2018}. First, we reconstruct 3D facial shapes using each of the compared methods from the 70 2D images in the test set. Then, for each of the estimated 3D faces, we compute reconstruction errors with respect to the ground truth 3D scan as follows: 1) using a set of facial landmarks (manually placed in the scans and in the reconstructions from \citep{BaiCVPR2020}, and derived from the corresponding 3DMM for \citep{JGuoECCV2020} and \citep{BasACCV2016}), we rigidly align the reconstructed 3D face to the ground truth 3D scan using Procrustes analysis; 2) we compute point-to-surface Euclidean distance from each vertex in the reconstructed 3D face to the 3D scan, and from the scan to the reconstruction; finally, 3) the reconstruction error of a 3D facial estimation is computed as the average between the mean of the two measures computed in the previous step. Nevertheless, notice that each reconstruction and 3D facial scan cover different facial areas, and thus the above procedure may lead to an unfair comparison. To allow for an adequate evaluation, we define a facial region common to both  reconstructions and scans using anatomical landmarks, and compute the errors only for the vertices in this facial region.

\subsection{Results}\label{subsec:Results}
\begin{figure}[htbp]
    \centering
    \includegraphics[width=0.6 \textwidth]{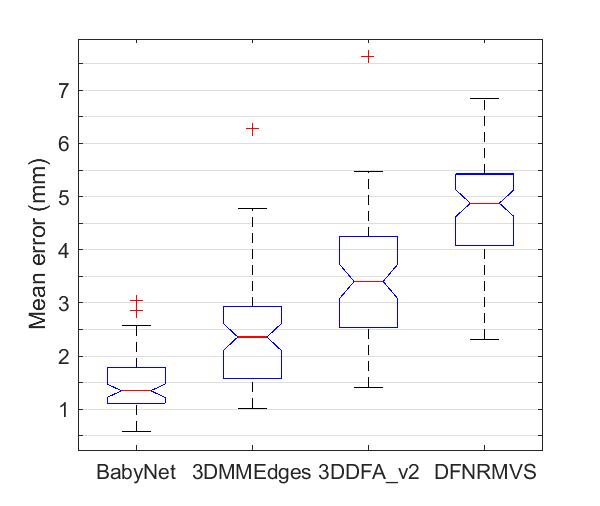}
    \caption{Reconstruction errors obtained with the compared methods. BabyNet is the proposed method, 3DMMEdges is the 3DMM-fitting method proposed by \citep{BasACCV2016}, 3DDFA\_V2 corresponds to the deep learning method proposed by \citep{JGuoECCV2020}, and DFNRMVS to the deep learning method proposed by \cite{BaiCVPR2020}.}
    \label{fig:Boxplots}
\end{figure}

Figure \ref{fig:Boxplots} shows in boxplots the reconstruction errors obtained with each of the compared methods. On the one hand, the proposed BabyNet clearly outperforms the deep learning methods trained with data from adults (3DDFA\_V2 \citep{JGuoECCV2020} and DFNRMVS \citep{BaiCVPR2020}), suggesting the need for a method specifically designed for baby data. On the other hand, although a baby 3DMM has been used to apply the 3DMM-fitting method \citep{BasACCV2016} (3DMMEdges), the resulting 3D facial reconstructions show higher errors than the BabyNet, proving the potential of deep learning methods to improve the quality of the 3D facial reconstructions.

\begin{figure}[htbp]
    \begin{subfigure}[t]{0.3\textwidth}
        \centering
        \includegraphics[width=\textwidth]{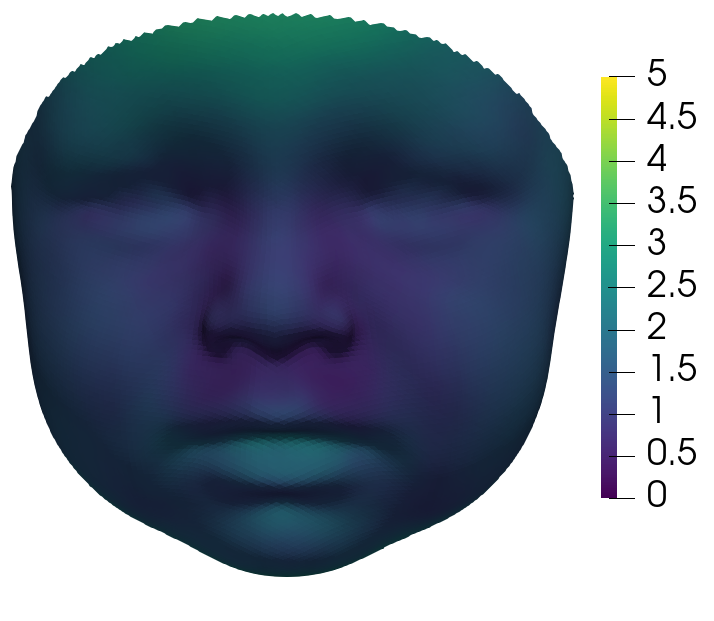}
        \caption{BabyNet (ours)}\label{fig:ErrorMap_BabyNet}
    \end{subfigure}
    \hfill
    \begin{subfigure}[t]{0.3\textwidth}
        \centering
        \includegraphics[width=\textwidth]{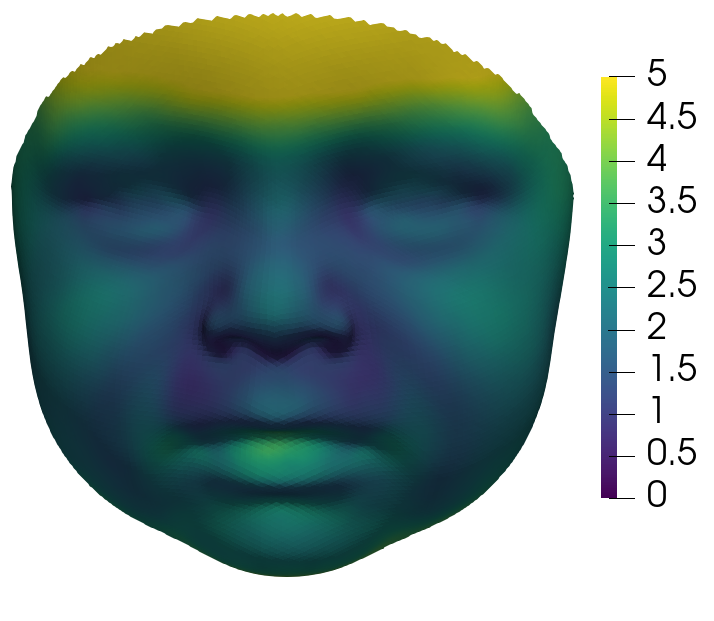}
        \caption{3DMMEdges \citep{BasACCV2016}}\label{fig:ErrorMap_Bas2016}
    \end{subfigure}
    \hfill
    \begin{subfigure}[t]{0.3\textwidth}
        \centering
        \includegraphics[width=\textwidth]{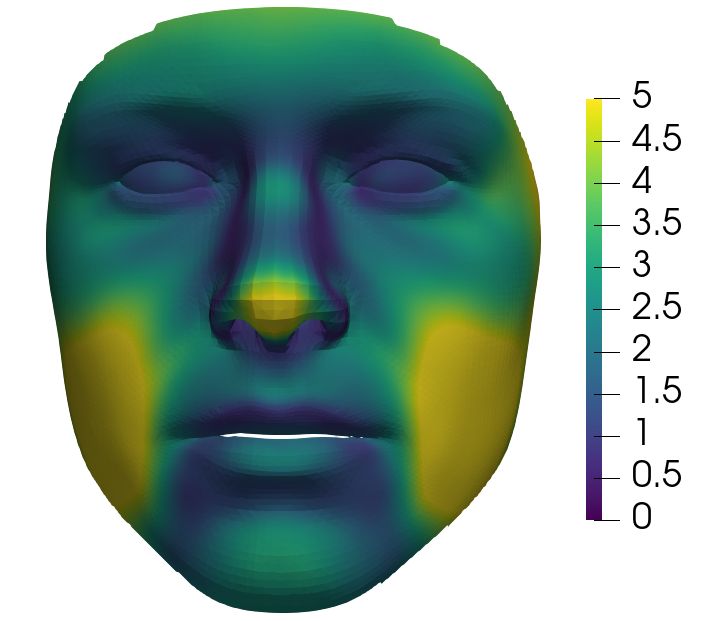}
        \caption{3DDFA\_V2 \citep{JGuoECCV2020}}\label{fig:ErrorMap_JGuo2020}
    \end{subfigure}
    \caption{Reconstruction errors in millimetres shown as colour maps. We only show the point-to-surface errors for the reconstruction to the original 3D facial scan since the scans do not have a common triangulation, and thus the mean error per vertex cannot be determined.}
    \label{fig:errorMaps}
\end{figure}

These results are complemented with Figure \ref{fig:errorMaps}, where we show the mean reconstruction errors per vertex as colour maps\footnote{DFNRMVS \cite{BaiCVPR2020} cannot be represented in Figure \ref{fig:errorMaps} since it does not keep the order of the correspondences. Nevertheless, its error is clearly above the rest of the other methods, as it can be seen in Figure \ref{fig:Boxplots}.}. Since the ground truth 3D facial scans do not have a common triangulation, the per-vertex mean reconstruction error cannot be determined. Therefore, we only show the point-to-surface distances from the estimated shapes to the scans. From Figure \ref{fig:errorMaps}, we observe that adult methods have higher errors in regions such as the cheeks or the nose, which are probably the areas most different between adults and babies. The highest errors in the forehead of the 3DMM-fitting method (3DMMEdges) can be explained by the absence of image edges in the forehead, since this method is based on the correspondence between image edges and occluding boundaries in 3D. Even so, the errors of the most inner part of the face are not as low as the obtained with the BabyNet.

\begin{figure}[htbp]
    \begin{subfigure}[t]{0.15\textwidth}
        \centering
        \includegraphics[width=\textwidth]{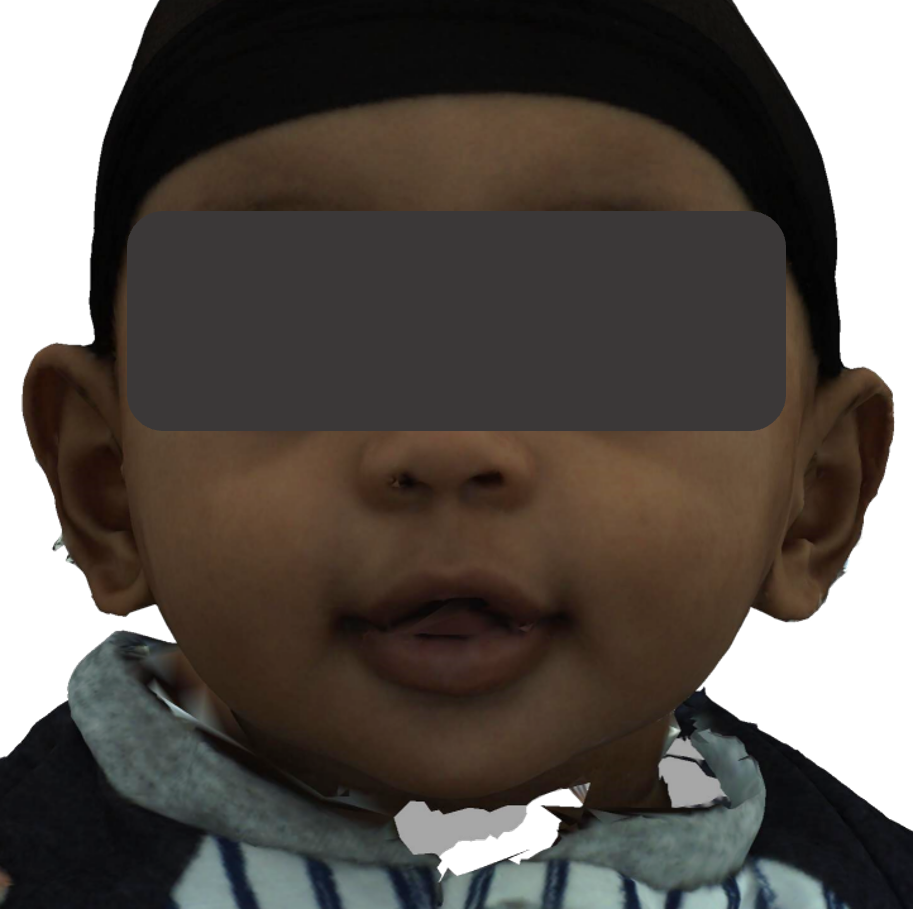}
    \end{subfigure}
    \begin{subfigure}[t]{0.15\textwidth}
        \centering
        \includegraphics[width=\textwidth]{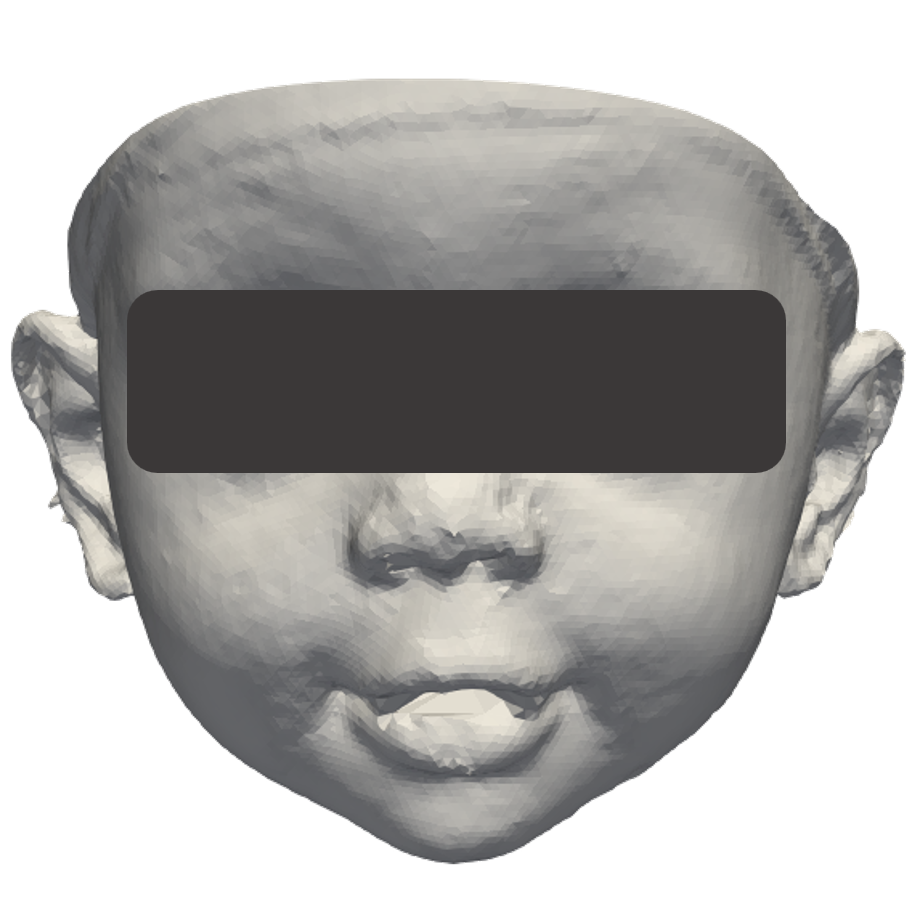}
    \end{subfigure}
    \begin{subfigure}[t]{0.15\textwidth}
        \centering
        \includegraphics[width=\textwidth]{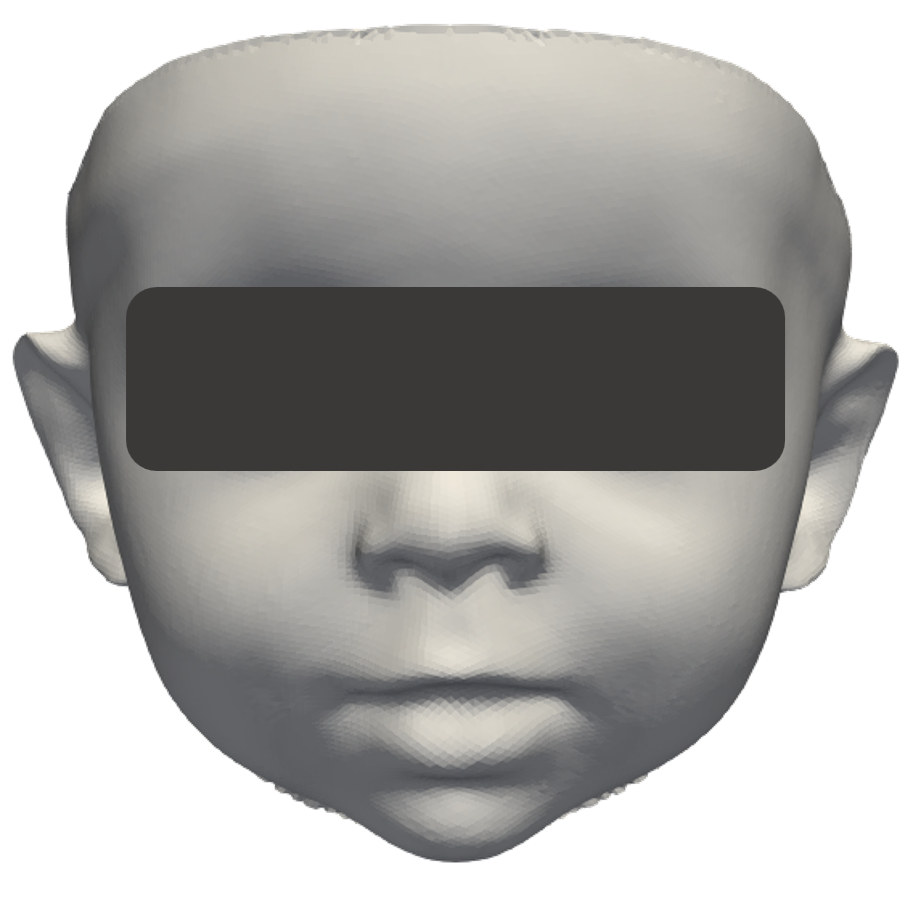}
    \end{subfigure}
    \hfill
    \begin{subfigure}[t]{0.15\textwidth}
        \centering
        \includegraphics[width=\textwidth]{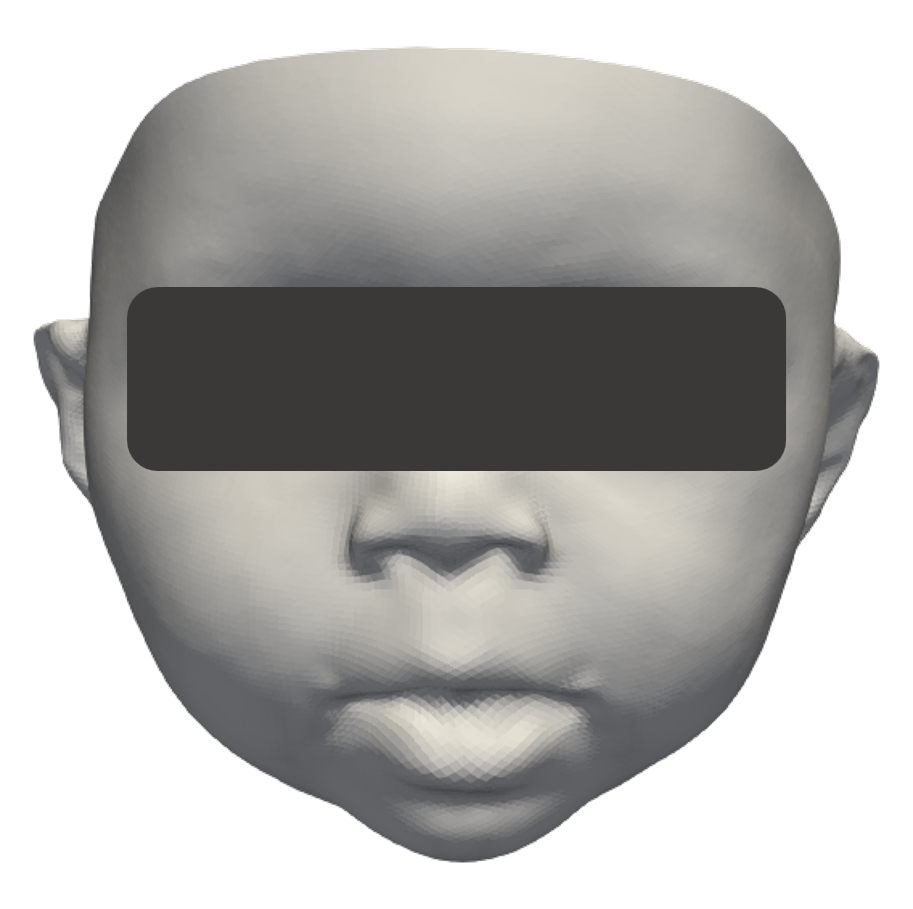}
    \end{subfigure}
    \hfill
    \begin{subfigure}[t]{0.15\textwidth}
        \centering
        \includegraphics[width=\textwidth]{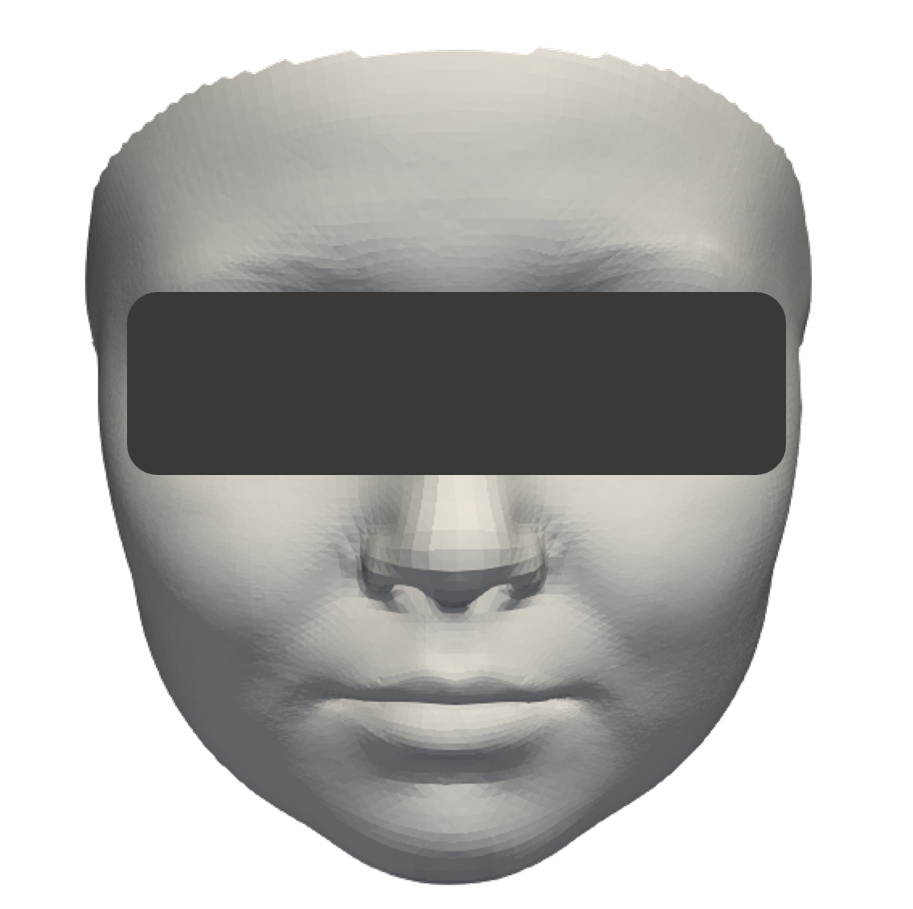}
    \end{subfigure}
    \hfill
    \begin{subfigure}[t]{0.15\textwidth}
        \centering
        \includegraphics[width=\textwidth]{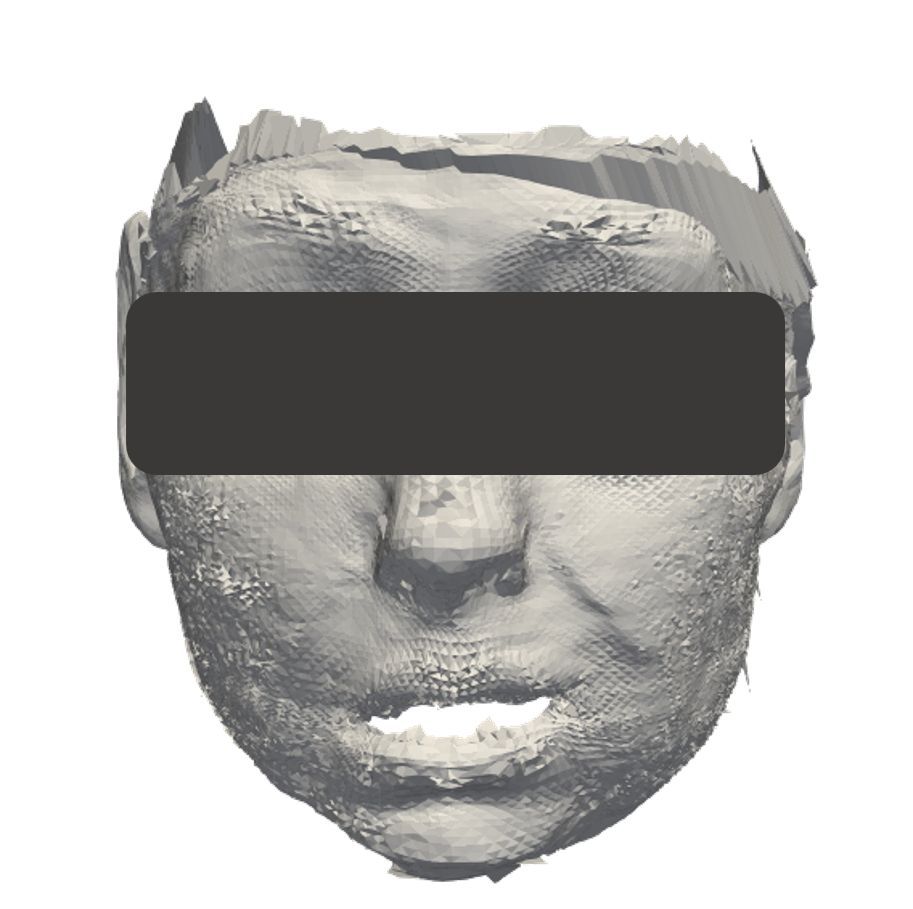}
    \end{subfigure}
    \hfill
    \begin{subfigure}[t]{0.15\textwidth}
        \centering
        \includegraphics[width=\textwidth]{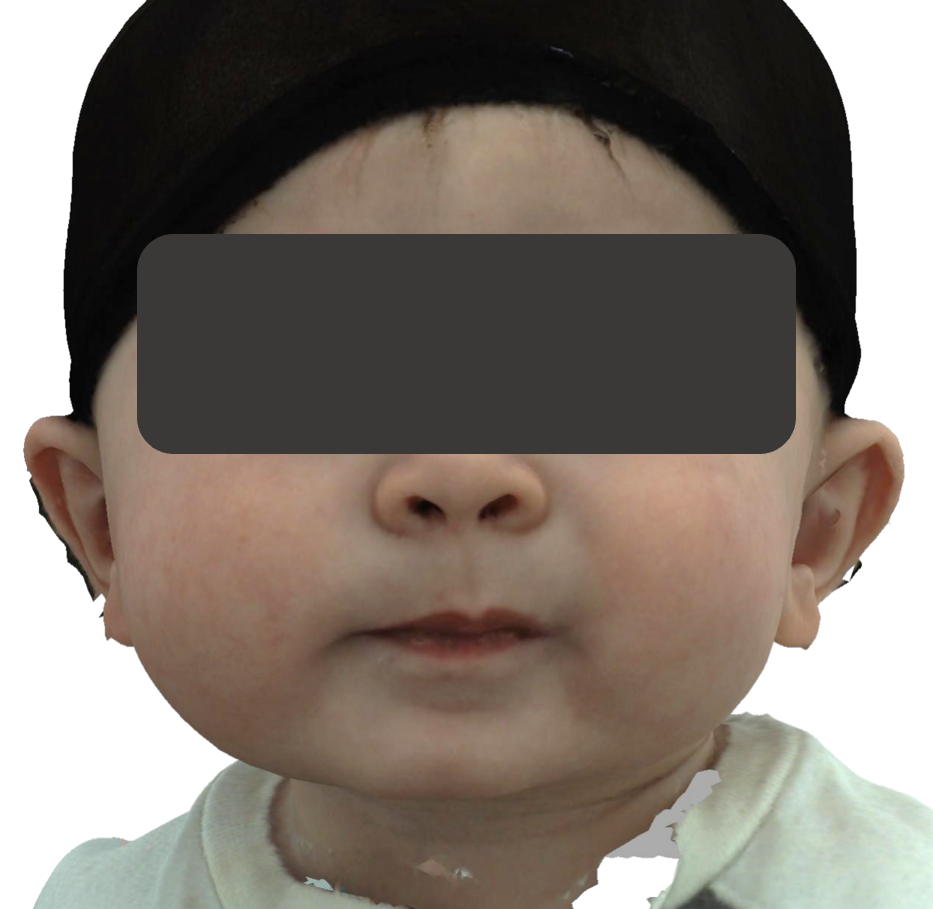}
    \end{subfigure}
    \begin{subfigure}[t]{0.15\textwidth}
        \centering
        \includegraphics[width=\textwidth]{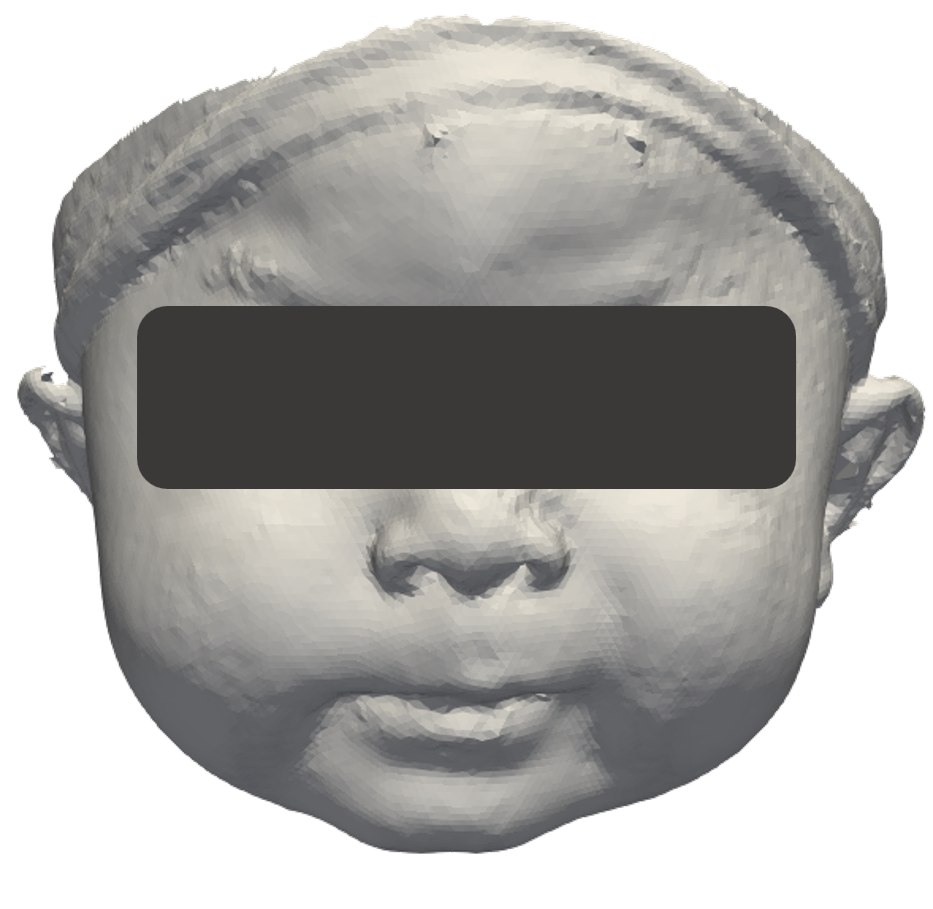}
    \end{subfigure}
    \begin{subfigure}[t]{0.15\textwidth}
        \centering
        \includegraphics[width=\textwidth]{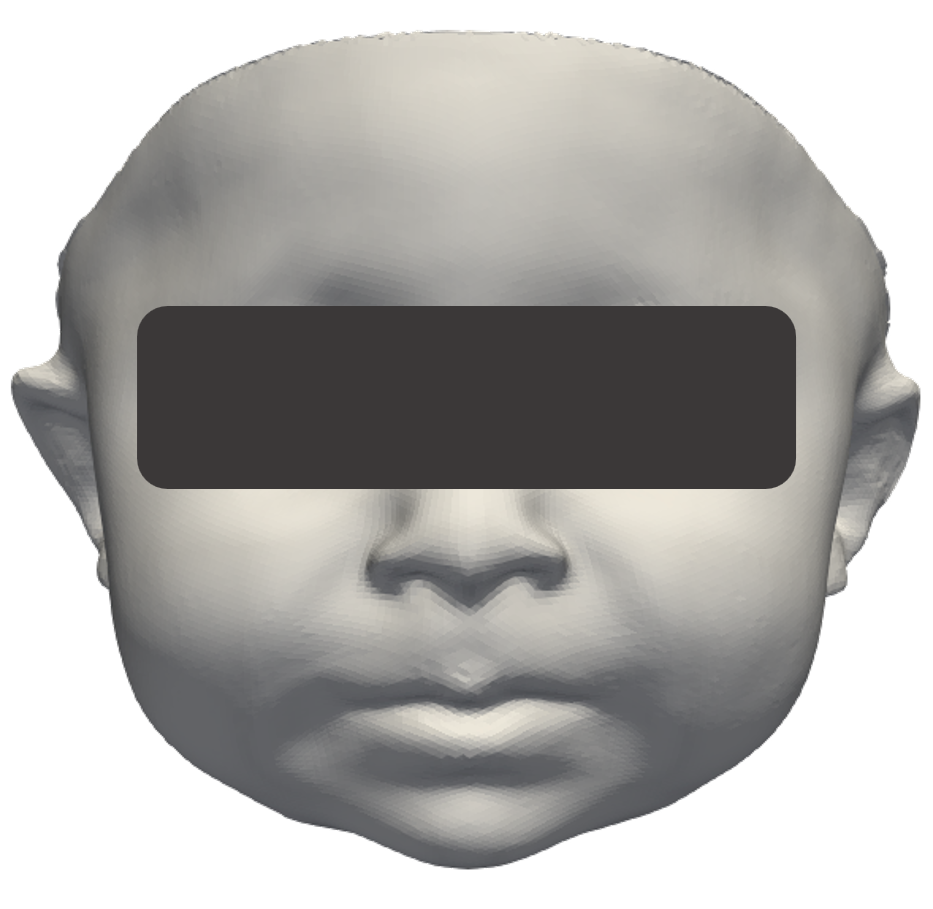}
    \end{subfigure}
    \hfill
    \begin{subfigure}[t]{0.15\textwidth}
        \centering
        \includegraphics[width=\textwidth]{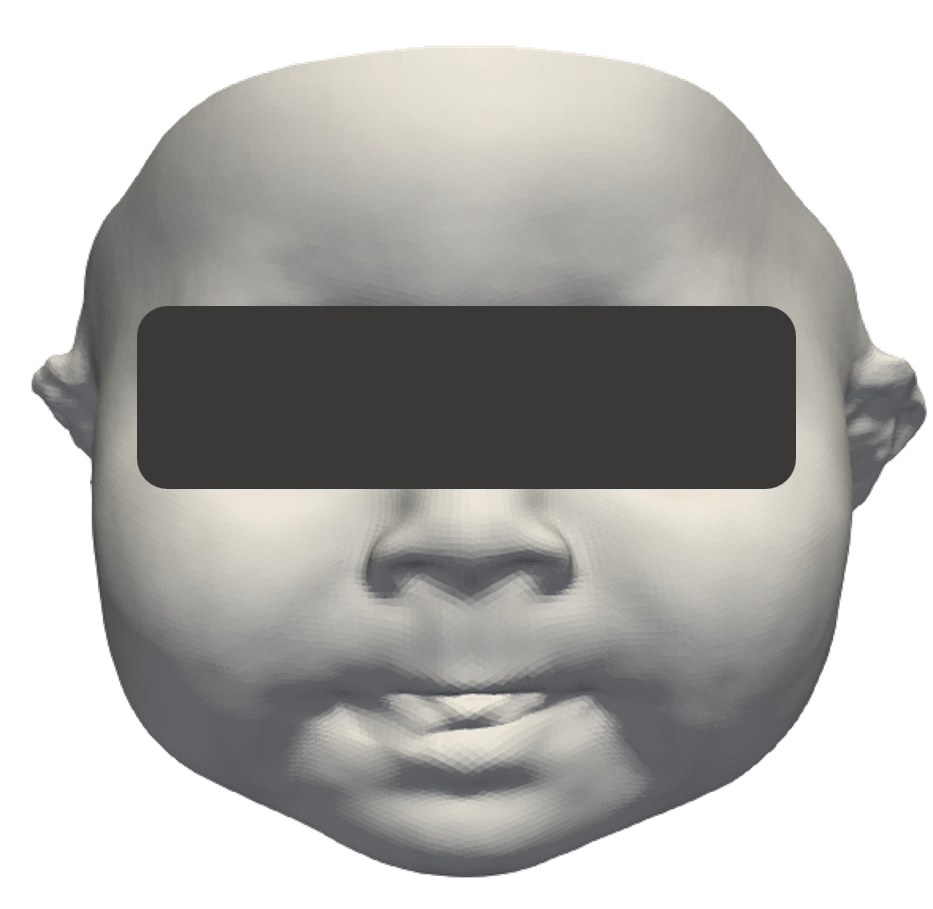}
    \end{subfigure}
    \hfill
    \begin{subfigure}[t]{0.15\textwidth}
        \centering
        \includegraphics[width=\textwidth]{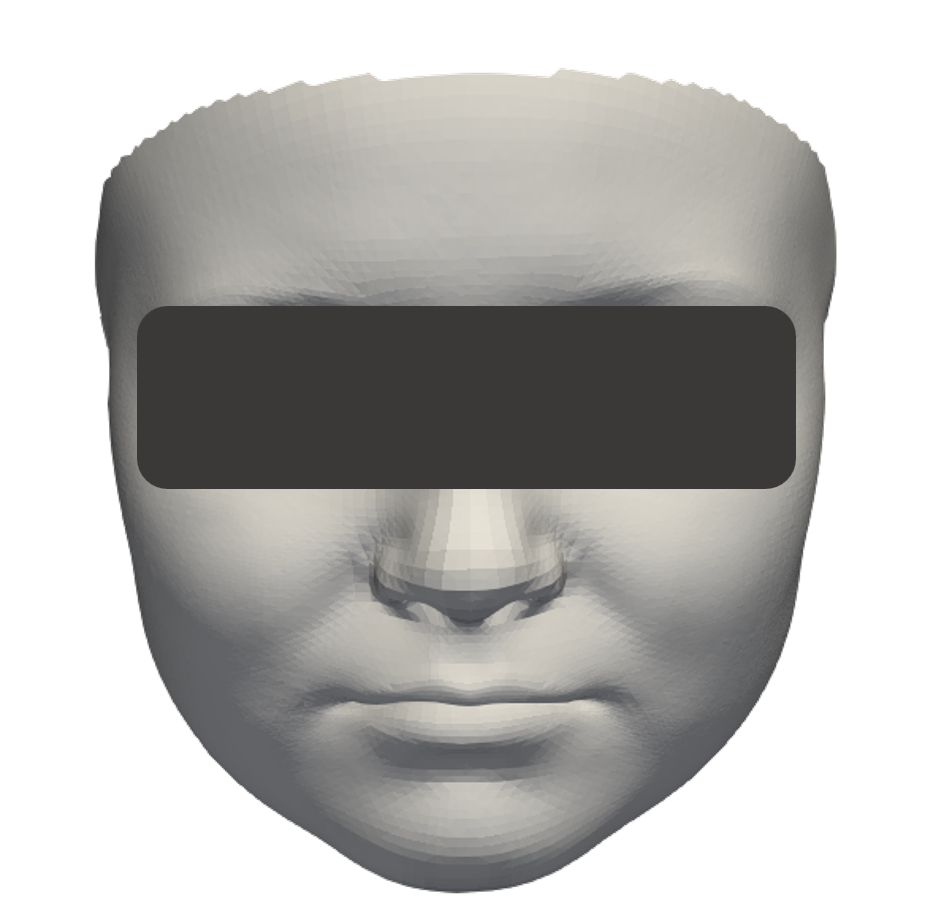}
    \end{subfigure}
    \hfill
    \begin{subfigure}[t]{0.15\textwidth}
        \centering
        \includegraphics[width=\textwidth]{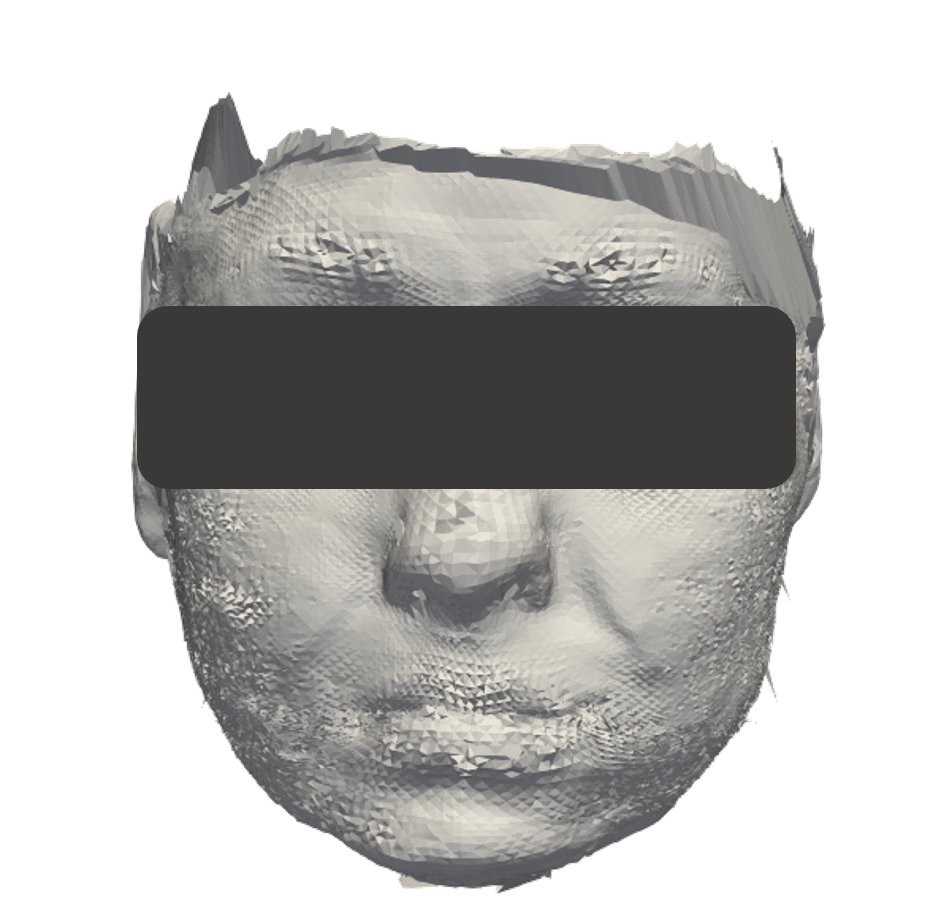}
    \end{subfigure}
    \hfill
    \begin{subfigure}[t]{0.15\textwidth}
        \centering
        \includegraphics[width=\textwidth]{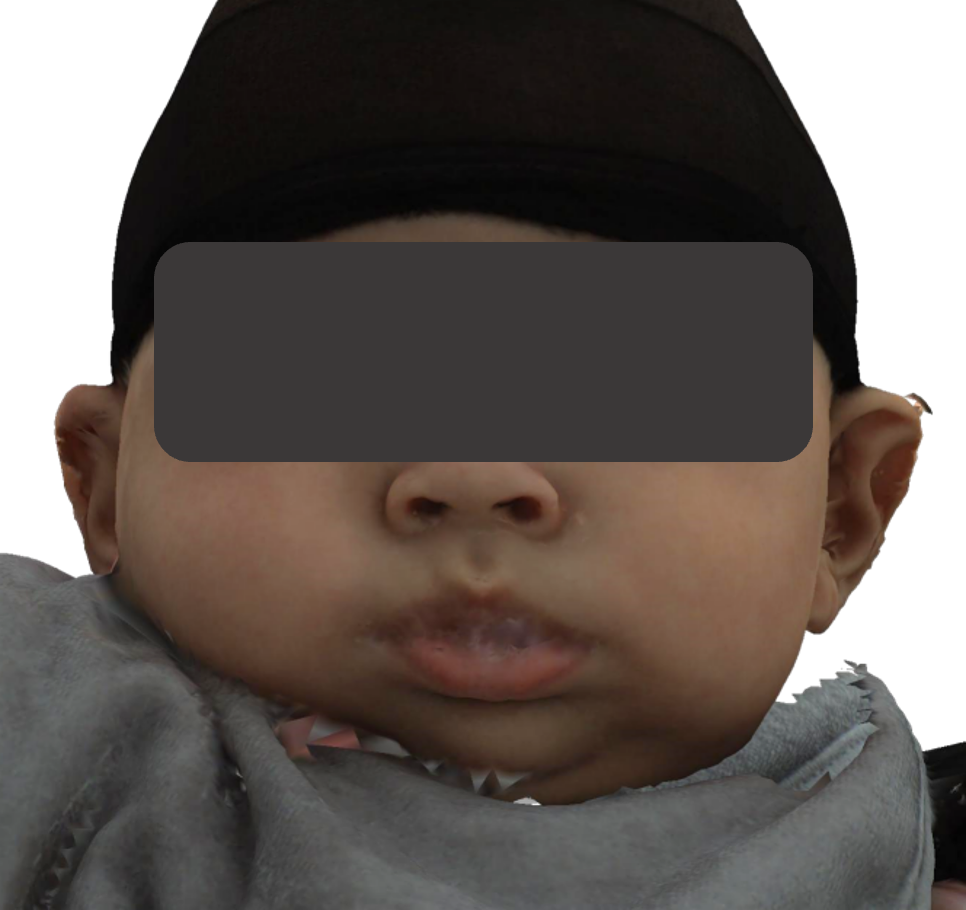}
        \caption{Image}
    \end{subfigure}
    \hfill
    \begin{subfigure}[t]{0.15\textwidth}
        \centering
        \includegraphics[width=\textwidth]{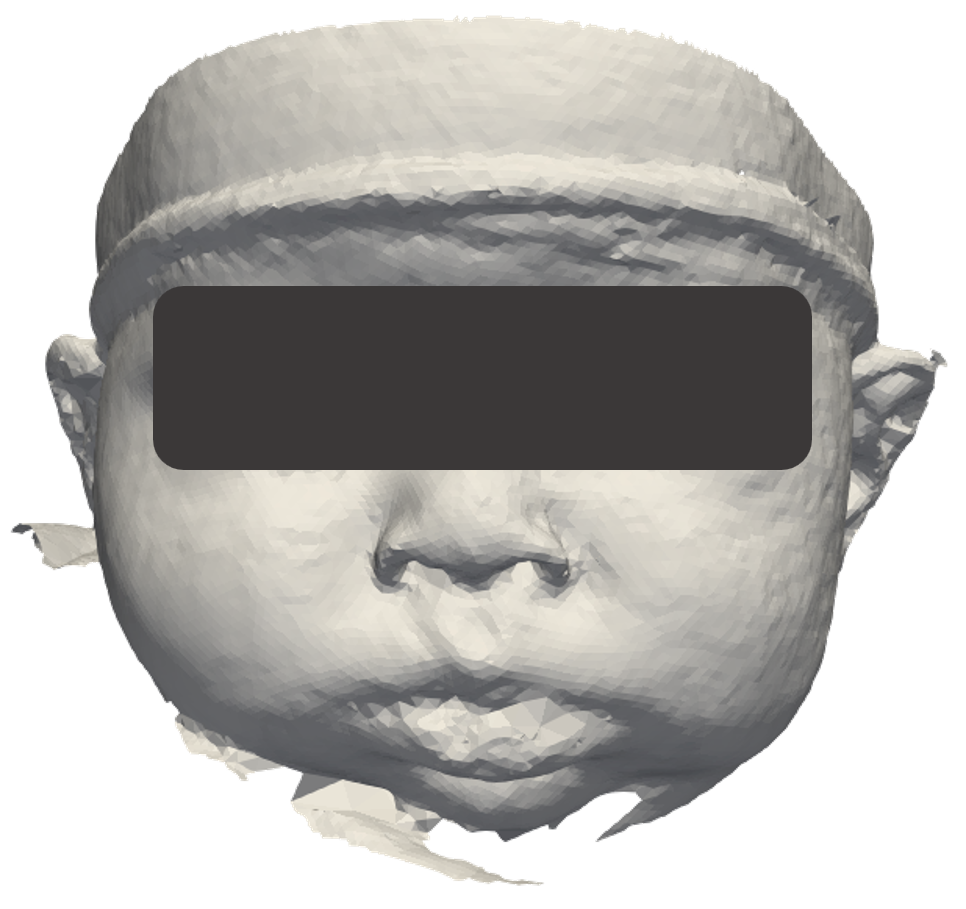}
        \caption{Original scan}
    \end{subfigure}
    \hfill
    \begin{subfigure}[t]{0.15\textwidth}
        \centering
        \includegraphics[width=\textwidth]{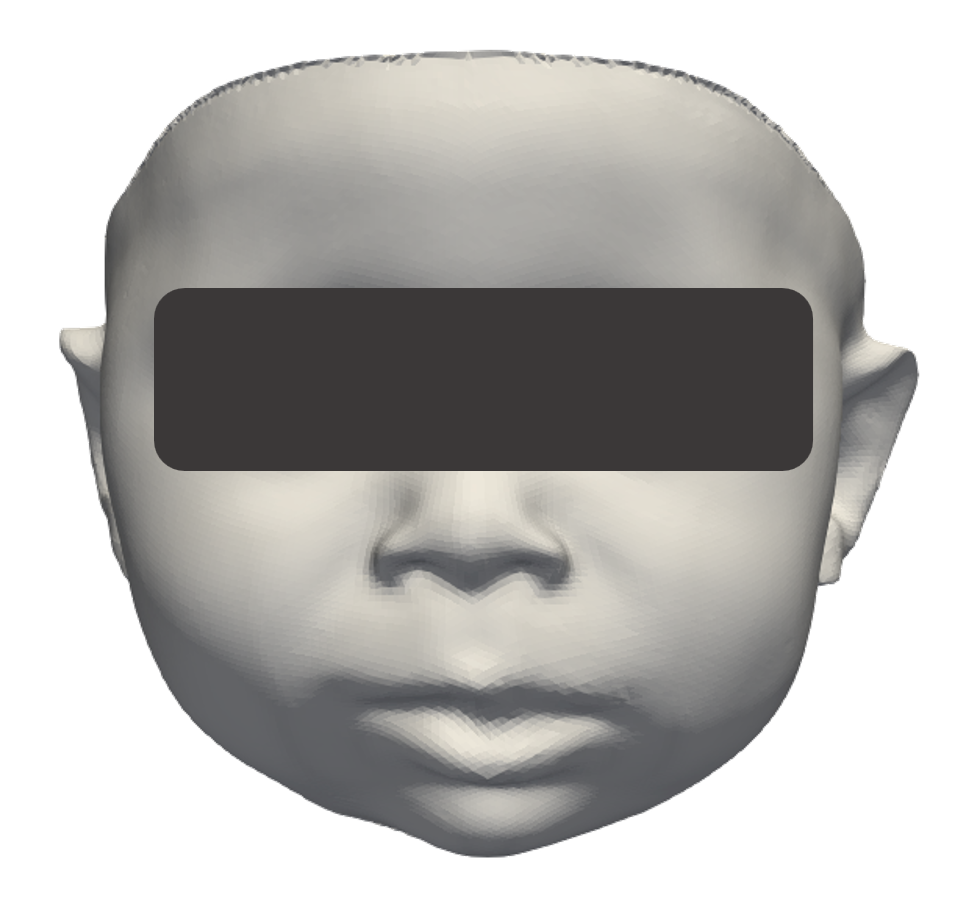}
        \caption{BabyNet (ours)}\label{fig:examples_3Drecs_BabyNet}
    \end{subfigure}
    \hfill
    \begin{subfigure}[t]{0.15\textwidth}
        \centering
        \includegraphics[width=\textwidth]{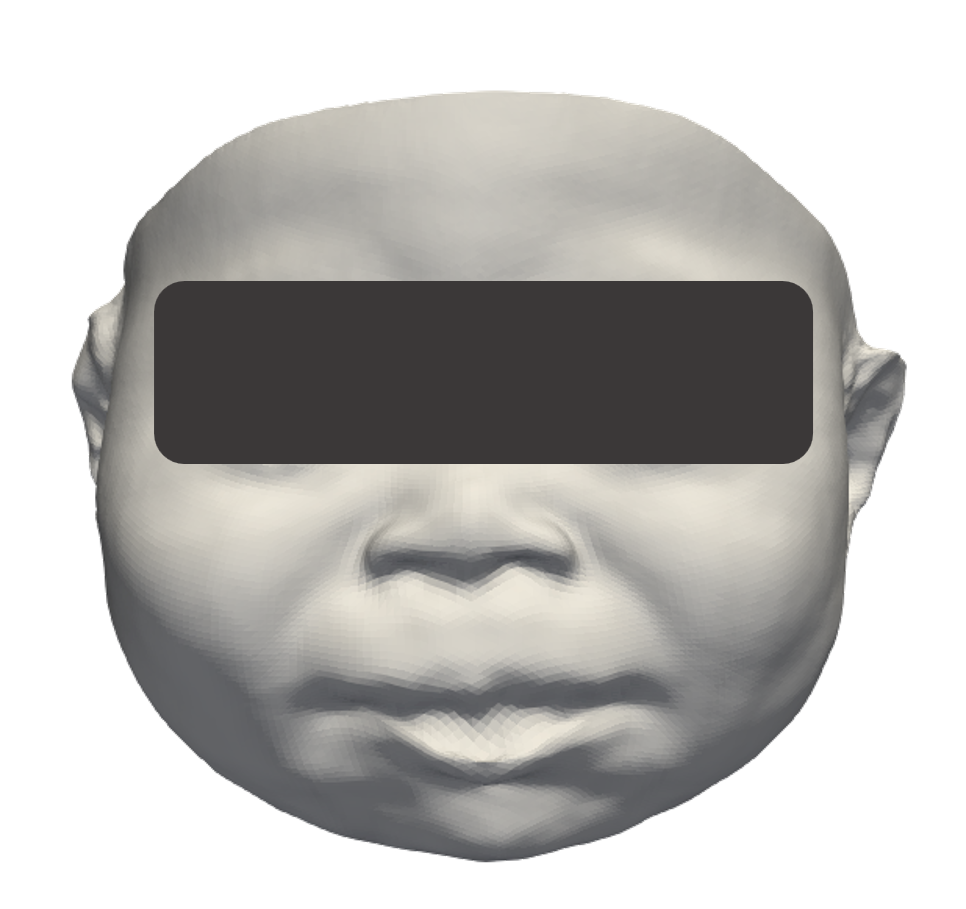}
        \caption{3DMMEdges \citep{BasACCV2016}}\label{fig:examples_3Drecs_Bas2016}
    \end{subfigure}
    \hfill
    \begin{subfigure}[t]{0.15\textwidth}
        \centering
        \includegraphics[width=\textwidth]{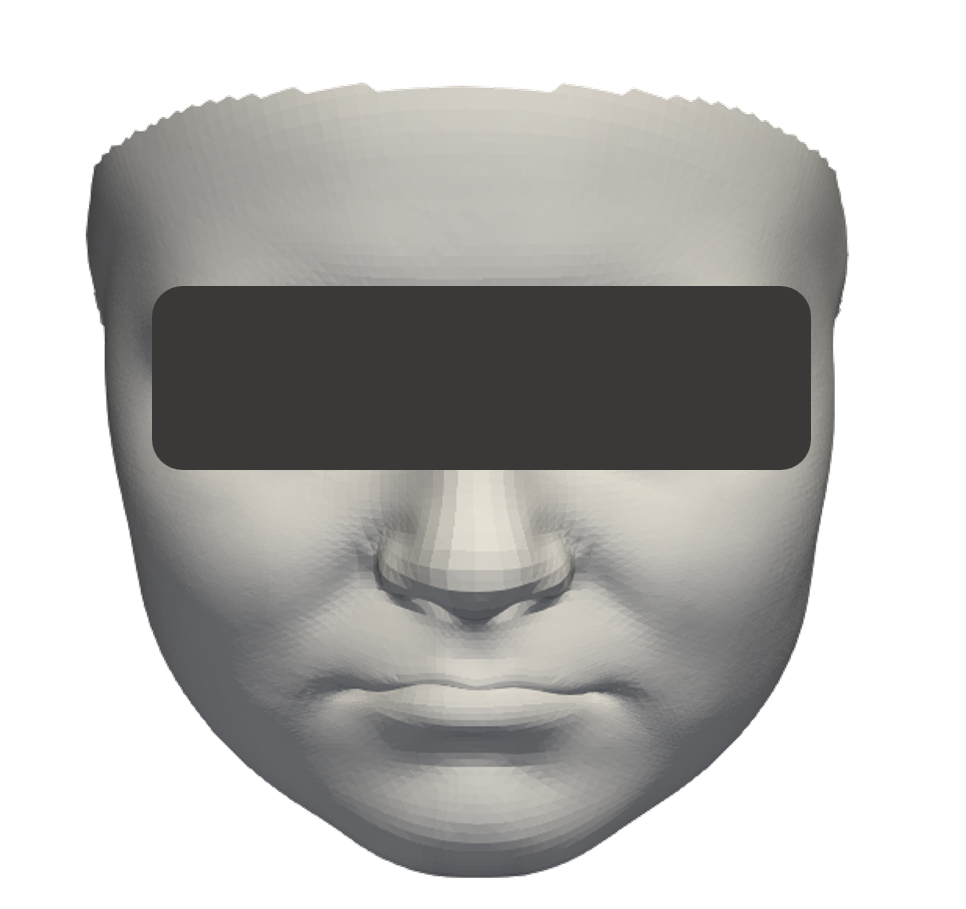}
        \caption{3DDFA\_V2 \citep{JGuoECCV2020}}\label{fig:examples_3Drecs_JGuo2020}
    \end{subfigure}
    \hfill
    \begin{subfigure}[t]{0.15\textwidth}
        \centering
        \includegraphics[width=\textwidth]{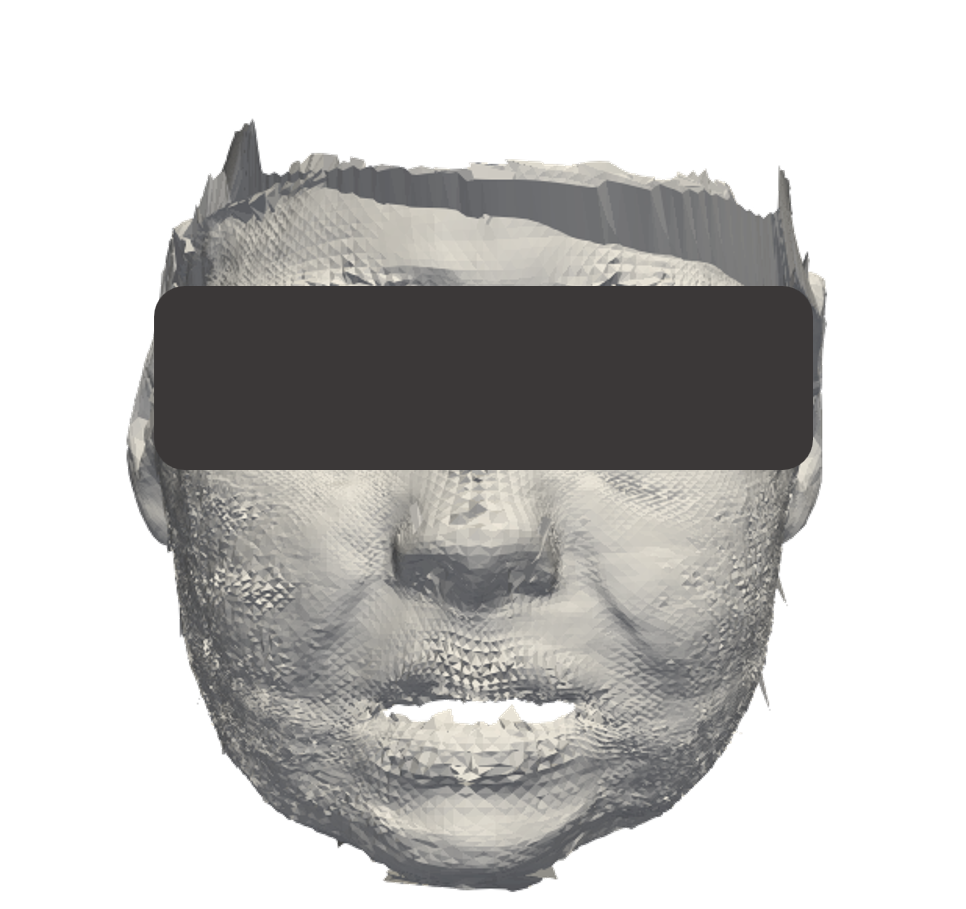}
        \caption{DFNRMVS \citep{BaiCVPR2020}}\label{fig:examples_3Drecs_Bai2020}
    \end{subfigure}
    \caption{Examples of the reconstructions obtained with the compared methods.}
    \label{fig:examples_3Drecs}
\end{figure}

Figure \ref{fig:examples_3Drecs} shows qualitatively the observations from above with examples of reconstructions obtained with the different methods. The BabyNet (Figure \ref{fig:examples_3Drecs_BabyNet}) is able to reconstruct the facial geometry of babies with higher visual accuracy than the other methods, recovering not only the global shape of the face, but also the expression and facial details. Again, the adult methods (Figures \ref{fig:examples_3Drecs_JGuo2020} and \ref{fig:examples_3Drecs_Bai2020}) are not able to capture the characteristic features of a baby, like chubby cheeks and rounded nose. On the other hand, the 3DMM-fitting method (Figure \ref{fig:examples_3Drecs_Bas2016}) does reconstruct 3D faces that look like babies, but they show artefacts due to the imprecise correspondences found between the image and the 3DMM.

\section{Discussion} \label{sec:7Discussions}
We presented the BabyNet, a 3D face reconstruction method designed to recover the 3D facial geometry of babies from uncalibrated 2D images. It consists of two consecutive phases: firstly, a 3D autoencoder is trained to estimate a representative non-linear low-dimensional latent space from which 3D faces can be recovered; secondly, the 3D encoder is replaced by a 2D encoder that extracts the same representative features from 2D images, hence the 3D decoder is able to recover the 3D facial shape from the feature vector estimated from the input image.

For the 3D autoencoder, we use an emerging type of neural networks that extent the successful convolutional neural networks to non-Euclidean data, the graph neural networks. Using this type of networks, we can directly process the 3D facial meshes, avoiding the need for intermediate Euclidean representations, which are often not as accurate, and thus limit the reconstruction process. On the other hand, the 2D encoder uses a powerful pre-trained network that extracts meaningful features from facial images. We extract features from both intermediate layers and the output layer, thus capturing features at different levels. The intermediate layers allow us to accurately recover not only the global shape of the face, but also details, as we show quantitatively and qualitatively. These image features are mapped to the latent space learnt by the 3D autoencoder so that the 3D decoder can reconstruct the corresponding 3D facial shape.

We quantitatively and qualitatively evaluate the proposed method by comparing its reconstruction accuracy to that of three state-of-the-art methods: two based on deep learning but trained with adult data \citep{JGuoECCV2020,BaiCVPR2020}, and another one based on 3DMM-fitting \citep{BasACCV2016}. We show that 1) although deep learning has proven to be successful for 3D face reconstruction, it has to be trained with baby data when targeting the reconstruction of infant faces, since methods trained with adult data are not able to represent the characteristic facial features of babies; and that 2) the reconstruction power of classical 3DMM-fitting methods is surpassed by the deep learning-based BabyNet, even when a baby-specific 3DMM, like the BabyFM, is used. These two observations show the value of the BabyNet: a 3D face reconstruction system specific to the 3D facial geometry of babies and based on deep learning so as to obtain more faithful 3D facial shapes.

Although our results outperform the state-of-the-art when targeting babies, our method also have limitations. First, given that the training data is fully synthetic, the reconstruction accuracy of the BabyNet over real 2D images may be reduced. Also, even though we are able to improve the reconstruction of details by using also intermediate feature maps from the 2D encoder, we are not able to reach the extend of details present in real 3D facial data since the training 3D data is synthetic. These remarks can be summarised into one limitation that is the reconstruction of details, which is also a current limitation of the 3D face reconstruction field. Besides, we believe that using a image feature extractor network trained specifically with photographs of babies would further improve the accuracy of the BabyNet, which may be also limited by the adult-trained ArcFace.

\section{Conclusion} \label{sec:8Conclusions}
Our BabyNet is the first 3D face reconstruction algorithm based on deep learning specifically trained to target babies. With the proposed architecture, we are able to reconstruct the global shape of the infant, but also finer features. We demonstrate that the BabyNet outperforms both complex deep learning-based 3D face reconstruction methods that have been trained with adult data, and classical model-fitting approaches, even with a baby 3D facial model is used.

\section*{Acknowledgements}
This work is partly supported by the Spanish Ministry of Science and Innovation under project grant PID2020-114083GB-I00 and the NIH Eunice Kennedy Shriver National Institute of Child Health \& Human Development grant R42 HD08171203.

\bibliography{journals-abbreviations,Bibliography_merged}

\end{document}